\documentstyle[12pt]{article}
\input epsf.sty

\newtheorem{Definition}{Definition}
\newtheorem{Theorem}{Theorem}
\newtheorem{Lemma}{Lemma}
\newtheorem{Corollary}{Corollary}
\newtheorem{Claim}{Claim}[Lemma]

\def\PFIN{{\rm PFIN}}
\def\FIN{{\rm FIN}}
\def\EX{{\rm EX}}

\newcommand{\AAA}{{\cal A}}
\newcommand{\AAB}{{\cal A}'}

\newcommand{\comment}[1]{{}}
\def\bbbn{{\rm I\!N}}
\def\bbbr{{\rm I\!R}}
\def\bbbq{{\mathchoice {\setbox0=\hbox{$\displaystyle\rm
Q$}\hbox{\raise
0.15\ht0\hbox to0pt{\kern0.4\wd0\vrule height0.8\ht0\hss}\box0}}
{\setbox0=\hbox{$\textstyle\rm Q$}\hbox{\raise
0.15\ht0\hbox to0pt{\kern0.4\wd0\vrule height0.8\ht0\hss}\box0}}
{\setbox0=\hbox{$\scriptstyle\rm Q$}\hbox{\raise
0.15\ht0\hbox to0pt{\kern0.4\wd0\vrule height0.7\ht0\hss}\box0}}
{\setbox0=\hbox{$\scriptscriptstyle\rm Q$}\hbox{\raise
0.15\ht0\hbox to0pt{\kern0.4\wd0\vrule height0.7\ht0\hss}\box0}}}}
\newcommand{\const}{\mathop{\rm const}\nolimits}
\begin{document}

\title{\bf Probabilistic and Team PFIN-type Learning: General Properties}

\author{Andris Ambainis\\
	Department of Combinatorics and Optimization \\
        and Institute for Quantum Computing\\
	University of Waterloo\\
        200 University Avenue West\\
	Waterloo, ON N2T 2L2, Canada\\
	e-mail:~ambainis@math.uwaterloo.ca}
\date{}

\maketitle

\begin{abstract}
We consider the probability hierarchy for Popperian FINite learning
and study the general properties of this hierarchy.
We prove that the probability hierarchy is decidable, i.e.
there exists an algorithm that receives $p_1$ and $p_2$ and
answers whether $\PFIN$-type learning with the probability of success $p_1$
is equivalent to $\PFIN$-type learning with the probability of success $p_2$.

To prove our result, we analyze the topological structure of the probability
hierarchy. We prove that it is well-ordered in descending ordering
and order-equivalent to ordinal $\epsilon_0$.
This shows that the structure of the hierarchy is very complicated.

Using similar methods, we also prove that, for $\PFIN$-type learning,
team learning and probabilistic learning are of the same power.
\end{abstract}

\section{Introduction}

Inductive inference is a branch of theoretical
computer science that studies the process of learning
in a recursion-theoretic framework \cite{Gold,AS,OSW}.
Within inductive inference,
there has been much work on team learning
(see surveys in \cite{Ambainis99,JS95,Smith94}).


Probabilistic learning is closely related to team learning.
Any team of machines can be simulated by a single probabilistic machine
with the same success ratio.
The simulation of a probabilistic machine by a team of deterministic
machines is often possible as well.

In this paper, we consider finite learning of total recursive
functions (abbreviated as FIN).
The object to be learned is a total recursive function $f$.
A learning machine reads the values of the function
$f(0)$, $f(1)$, $\ldots$ and produces
a program computing $f$ after having seen a finite initial
segment of $f$.
The learning machine is not allowed to change the program later.

$\FIN$ is supposed to be one of the simplest learning para\-digms.
However, if we consider probabilistic and team learning,
the situation becomes very complex.
Probabilistic $\FIN$-type learning has been
studied since 1979 but we are still far from the complete understanding
of this area.

The investigation of probabilistic FINite learning
was started by Freivalds\cite{Freivalds}.
He gave a complete description of the learning capabilities
for probabilistic machines with probabilities of success
above $\frac{1}{2}$.
These results were extended to team learning by 
Daley et. al.\cite{Vel,DKV91}.

The further progress was very difficult.
Daley, Kalyanasundaram and Velauthapillai\cite{DKV92}
determined the capabilities for probabilistic learners 
with success probabilities in the interval
$[\frac{24}{49}, \frac{1}{2}]$.
Later, Daley and Kalyanasunda\-ram\cite{DK97} 
extended that to the interval $[\frac{12}{25}, \frac{1}{2}]$.
Proofs became more and more complicated.
(The full version of \cite{DK97} is more
than 100 pages long.)

$\PFIN$ (Popperian $\FIN$)-type learning 
is a simplified version of $\FIN$-type learning. 
In a $\PFIN$-type learning, a learning machine is allowed 
to output only programs computing total recursive functions.
Many properties of probabilistic and team $\PFIN$-type learning 
are similar to $\FIN$-type learning.
Yet, $\PFIN$-type learning is simpler and easier to analyse 
than unrestricted $\FIN$-type learning.

Daley, Kalyanasundaram and Velauthapillai\cite{DK93,DKV92a}
determined the capabilities of probabilistic 
$\PFIN$-type learners in the interval $[\frac{3}{7}, \frac{1}{2}]$.
However, even for $\PFIN$-type learning, the situation
becomes more and more complicated for smaller probabilities of success.

In this paper, we suggest another approach to $\PFIN$-type and $\FIN$-type
learning.
Instead of trying to determine the exact points at which the learning
capabilities are different (either single points or sequences
of points generated by a formula),
we investigate global properties of the probability structure.

Our main result is that the probability hierarchy for $\PFIN$-type
learning 
is well-ordered in a decreasing ordering and has a constructive description
similar to systems of notations for constructive ordinals.
We use this result to construct a decision algorithm for 
the probability hierarchy.
Given two numbers $p_1, p_2\in[0, 1]$, the decision algorithm answers
whether the learning with probability $p_1$ is equivalent 
to the learning with probability $p_2$.
Also, we construct a universal simulation algorithm
receiving 
\begin{itemize}
\item
$p_1, p_2\in[0, 1]$ such that $\PFIN$-learning
with these probabilities is equivalent and 
\item
$\PFIN$-learning machine $M$ with the probability of success $p_1$
\end{itemize}
and transforming $M$ into machine $M'$ with the probability of
success $p_2$.

All of these results make heavy use of the 
well-ordering and the system of notations.
To our knowledge, this is the first application
of well-orderings to a problem of this character.
(They have been used in computational learning 
theory\cite{Ambainis,FS93}, but for entirely different
purposes.)

We also determine the exact ordering type of the probability hierarchy.
It is order-isomorphic to $\epsilon_0$,
a quite large ordinal.\footnote{It is known that $\epsilon_0$ is isomorphic 
to the set of all expressions possible in first-order arithmetic.}
The part of the hierarchy investigated before ($[\frac{3}{7},1]$)
is order-isomorphic to the ordinal $3\omega$ and
is very simple compared to the entire probability hierarchy.
Thus, we can conclude that finding a more explicit  
description for the whole hierarchy is unlikely.
(The previous research shows that, even for segments
like $[\frac{3}{7}, 1]$ with a simple topological structure,
this task is difficult because of irregularities in
the hierarchy\cite{DK93}.)

Our results also imply that any probabilistic 
$\PFIN$-type learning machine can
be simulated by a team of deterministic machines with the same 
success ratio.

\section{Technical preliminaries}
\subsection{Notations}

We use the standard recursion theoretic notation\cite{Rogers}.

$\bbbn$ denotes $\{0, 1, \ldots\}$, the set of natural numbers.
$\bbbn^{+}$ denotes $\{1, 2, \ldots\}$,
the set of positive natural numbers,
$\bbbq$ denotes the set of rational numbers
and $\bbbr$ the set of real numbers. 
$\subseteq$ and $\subset$ denote a subset and 
a proper subset, respectively.

Let $\varphi$ denote an arbitrary fixed {\em acceptable programming system} 
(a.k.a. {\em G\"{o}del numbering}) of
all partial recursive functions\cite{Rogers1,Rogers,MY}.
$\varphi_i$ denotes the $i^{\rm th}$ program in system $\varphi$.

\subsection{Finite learning of functions}

A {\em learning machine} is an algorithmic device which reads
values of a recursive function $f$: $f(0)$, $f(1)$, $\ldots$.
Having seen finitely many values of the function
it can output a conjecture.
{\em The conjecture} is a program in some fixed acceptable 
programming system.
Only one conjecture is allowed, i.e.
the learning machine cannot change its conjecture later.

A learning machine $M$ {\em $\FIN$-learns} a function $f$ if,
receiving $f(0), f(1), \ldots$ 
as the input, it produces a program computing $f$.
$M$ $\FIN$-learns a set of functions $U$ if
it $\FIN$-learns any $f\in U$.
A set of functions $U$ is {\em $\FIN$-learnable}
if there exists a learning machine that learns $U$.
The collection of all $\FIN$-learnable sets is denoted $\FIN$.

$\PFIN$-learning is a restricted form of $\FIN$-learning.
A learning machine $M$ {\em $\PFIN$-learns} $U$ 
if it $\FIN$-learns $U$ and all conjectures (even incorrect ones)
of $M$ on all inputs are programs computing total recursive functions.
The collection of all $\PFIN$-learnable sets is denoted PFIN.

\subsection{Probabilistic and team learning}

Scientific discoveries are rarely done by one person.
Usually, a discovery is the result of collective effort.
In the area of computational learning theory, this observation
has inspired the research on team learning.

A team is just a set of learning machines:
$M=\{ M_1, \ldots, M_s \}$.
The team $M$ $[r, s]\FIN$-learns a set of functions $U$
if, for every $f\in U$, at least 
$r$ of $M_1$, $\ldots$, $M_s$ $\FIN$-learn $f$.
The collection of all $[r, s]\FIN$-learnable
sets is denoted $[r, s]\FIN$.

We also consider learning by probabilistic machines.
A probabilistic machine has an access to a fair coin and
its output depends on both input and the outcomes of coin flips.

Let $M$ be a probabilistic learning machine.
$M$ \FIN$\langle p\rangle$-learns
($\FIN$-learns with probability $p$) a set of functions $U$ if, 
for any function $f\in U$, the probability that $M$ outputs 
a program computing $f$, given $f(0)$, $f(1)$, $\ldots$ as the input,
is at least $p$.
$\FIN\langle p \rangle$ denotes the collection of all
$\FIN\langle p \rangle$-learnable sets.
   
Probabilistic and team $\PFIN$-learning is defined 
by adding a requirement that all conjectures output by 
the probabilistic machine or any machine in the team must be programs 
computing total recursive functions.

\begin{Definition}
The probability hierarchy for $\FIN$ 
is the set $A\subseteq \bbbr\cap[0,1]$ such that
\begin{enumerate}
\item
For any two different $p_1, p_2\in A$,
\[ \FIN\langle p_1\rangle\neq \FIN\langle p_2\rangle\]
i.e., learning with probability of success $p_1$ is
not equivalent to learning with probability of success $p_2$.
\item
If $x\in A$, $x\leq p$ and $[x, p[$ does not contain any points 
belonging to $A$, then 
\[ \FIN\langle x\rangle= \FIN\langle p\rangle.\]
\end{enumerate}
\end{Definition}

Essentially, the probability hierarchy is the set of
those probabilities at which the learning capabilities of
probabilistic machines are different.

The probability hierarchy for $\PFIN$ is defined similarly.

\subsection{Well-orderings and ordinals}
\label{sub:orddef}

A linear ordering is a {\em well-ordering} if it does not contain
infinite descending sequences.
{\em Ordinals}\cite{Sierpinski} are standard representations
of well-orderings.

The ordinal 0 represents the ordering type of the empty set,
the ordinal 1 represents the ordering type of any 1 element set,
the ordinal 2 represents the ordering type of any 2 element set and so on.
The ordinal $\omega$ represents the ordering type
of the set $\{ 0, 1, 2, \ldots\}$.
The ordinal $\omega+1$ represents the ordering type 
of $\{ 0, 1, 2, \ldots\}$ followed by an element $\omega$.
The ordinal $2\omega$ represents the ordering type
$\{0, 1, 2, \ldots\}$ followed by 
$\{\omega, \omega+1, \omega+2, \ldots\}$.
Greater ordinals can be defined similarly \cite{Sierpinski}.
We use arithmetic operations on ordinals defined
in two different ways.

\begin{Definition}
\cite{Kuratowski}
\label{def:ordaritm}
Let $A$ and $B$ be two disjoint sets,
$\alpha$ be the ordering type of $A$ and
$\beta$ be the ordering type of $B$.
\begin{enumerate}
\item
$\alpha+\beta$ is the ordering type of $A\cup B$
ordered so that $x<y$ for any $x\in A$, $y\in B$
and order is the same within $A$ and $B$.
\item
$\alpha\beta$ is the ordering type of $A\times B$
ordered so that $(x_1, y_1)< (x_2, y_2)$
iff $x_1<x_2$ or $x_1=x_2$ and $y_1<y_2$.
\end{enumerate}
\end{Definition}

We note that both the sum and the product of ordinals 
are non-commutative.
For example, $1+\omega=\omega\neq \omega+1$.

\begin{Definition}
\cite{Kuratowski}
\label{def:substr}
$\alpha-\beta$ (the difference of $\alpha$ and $\beta$)
is an ordinal $\gamma$ such that $\alpha=\beta+\gamma$.
\end{Definition}

$\alpha-\beta$ always exists and is unique\cite{Kuratowski}.
We also use the natural sum and the natural product of ordinals.
These operations use the representation of ordinals as
exponential polynomials.
In this paper, we consider only ordinals which are less than or equal to
\[ \epsilon_0=\lim (\omega, \omega^{\omega}, \omega^{\omega^{\omega}},
  \omega^{\omega^{\omega^{\omega}}}, \ldots ) .\]
Any ordinal $\alpha<\epsilon_0$ can be uniquely expressed in the form
\[ \alpha=c_1\omega^{\alpha_1}+\ldots+c_n\omega^{\alpha_n}\]
where $\alpha_1>\alpha_2> \ldots >\alpha_n$ are smaller ordinals and
$c_1$, $c_2$, $\ldots$, $c_n\in\bbbn$.

\begin{Definition}
\cite{Kuratowski}
\label{def:ordaritm1}
Let 
\[ \alpha=c_1\omega^{\alpha_1}+\ldots+c_n\omega^{\alpha_n}\]
\[ \beta=d_1\omega^{\alpha_1}+\ldots+d_n\omega^{\alpha_n}\]
\begin{enumerate}
\item
The natural sum of $\alpha$ and $\beta$ is
\[ \alpha(+)\beta=(c_1+d_1)\omega^{\alpha_1}+
\ldots+(c_n+d_n)\omega^{\alpha_n}.\]
\item
$\alpha(\cdot)\beta$, the natural product 
of $\alpha$ and $\beta$ is the product of
base $\omega$ representations as polynomials.
$\omega^{\alpha_i}(\cdot)\omega^{\alpha_j}=
\omega^{\alpha_i(+)\alpha_j}$ and
$\alpha(\cdot)\beta$ is the natural sum
of $c_i d_j \omega^{\alpha_i (+) \alpha_j}$
for all $i, j$.
\end{enumerate}
\end{Definition}
Natural sum and natural product are commutative.
They can be used to bound the ordering type of unions.
\begin{Theorem}
\label{lem:ord1}
Let $A_1, \ldots, A_s$ be arbitrary subsets of a well-ordered set $A$,
$\alpha_1$, $\ldots$, $\alpha_s$ be the ordering types of $A_1$,
$\ldots$, $A_s$ and $\alpha$ be the ordering type of
$A_1\cup \ldots\cup A_s$. Then,
\[ \alpha\leq\alpha_1(+)\alpha_2 (+)\ldots(+) \alpha_s .\]
\end{Theorem}
The difference between this theorem and Definition \ref{def:ordaritm}
is that Definition \ref{def:ordaritm} requires $x<y$ for all
$x\in A$, $y\in B$ but Theorem \ref{lem:ord1} has no such requirement.
Next, we give a similar result for the natural product.

\begin{Theorem}
\label{lem:ord2}
Let $A_1, \ldots, A_s$ and $A$ be well-ordered sets with
ordering types $\alpha_1$, $\ldots$, $\alpha_s$ and $\alpha$,
respectively.
Assume that $f:A_1\times A_2\times\ldots\times A_s\rightarrow A$ is  
a strictly increasing function onto $A$, i.e.
\[ f(\alpha_1, \ldots, \alpha_{i-1}, \alpha_i, \alpha_{i+1}, \ldots, \alpha_s)<
f(\alpha_1, \ldots, \alpha_{i-1}, \alpha'_i, \alpha_{i+1}, \ldots, \alpha_s)\]
for all $i\in\{1, \ldots, s\}$ and $\alpha_i<\alpha'_i$.
Then
\[ \alpha\leq\alpha_1(\cdot)\alpha_2 (\cdot)\ldots(\cdot) \alpha_s .\]
\end{Theorem}

Both Theorem \ref{lem:ord1} and \ref{lem:ord2} will be used in
section \ref{Relative}.
We will also use the {\em transfinite induction},
a generalization of the usual mathematical induction.

\begin{Theorem}
\cite[Principle of transfinite induction]{Kuratowski}
Let $A$ be a well-ordered set and $P(x)$ be a predicate.
If
\begin{enumerate}
\item
$P(x)$ is true when $x$ is the smallest element of $A$, and
\item
$P(y)$ for all $y\in A$ which are smaller than $x$ implies $P(x)$,
\end{enumerate}
then $P(x)$ for all $x\in A$.
\end{Theorem}

\subsection{Systems of notations}
\label{sub:sysdef}

In this paper we use subsets of $\bbbq\cap[0, 1]$ that are 
well-ordered in decreasing ordering.
A subset of $\bbbq$ is well-ordered in decreasing ordering
if it does not contain an infinite monotonically increasing
sequence.

Church and Kleene\cite{CK,Kleene} introduced 
{\em systems of notations} for constructive ordinals.
Intuitively, a system of notations is a way of assigning
notations to ordinals which satisfies certain constraints
and allows to extract certain information about the ordinal 
from its notation.
Below, we adapt the definition by 
Church and Kleene \cite{CK,Kleene} to
well-ordered subsets of $\bbbq$.

Let $A$ be a subset of $\bbbq$ which is 
well-ordered in decreasing ordering.
All elements of $A$ can be classified as follows:
\begin{enumerate}
\item
The greatest element of the set $A$.
We call it the {\em maximal} element.
\item
Elements $x$ which have an immediately preceding
element in decreasing ordering
(i.e. an element $y$ such that $x<y$ and
$[x, y]$ does not contain any points belonging to $A$).
Such elements are called {\em successor} elements.
\item
All other elements $x\in A$.
They are called {\em limit} elements.
\end{enumerate}

\begin{Definition}                                         
A system of notations for $A$ is a tuple of functions
$\langle k_S, p_S, q_S \rangle : \bbbq\rightarrow\bbbn$
such that
\begin{enumerate}
\item
$k_S(x)$ is equal to
\begin{enumerate}
\item
0, if $x$ is the maximal element;
\item
1, if $x$ is a successor element;
\item
2, if $x$ is a limit element;
\item
3, if $x\notin A$;
\end{enumerate}
\item
If $k_S(x)=1$, then $p_S(x)$ is defined and
it is the element immediately preceding $x$ in
descending ordering.
\item
If $k_S(x)=2$, then $q_S(x)$ is defined
and it is a program computing a monotonically decreasing 
sequence of elements of the set $A$ converging to $x$.
\end{enumerate}
\end{Definition}

Systems of notations are convenient for manipulating
well-ordered sets in our proofs.
Possibly, a system of notation
is the most appropriate way of describing
the probability hierarchy for $\PFIN$.
The structure of this hierarchy is 
quite complicated (Section \ref{Relative}) 
and it seems unlikely that more explicit descriptions exist.
 
Below, we give a useful property of systems of notations.

\begin{Lemma}
\label{FindInterval}
Let $A\subseteq\bbbq$ be a set which is well-ordered in
descending ordering and has a system of notations $S$.
Let $f_1(p)$ be the largest number in $A$ such that $f_1(p)\leq p$
and $f_2(p)$ be the smallest number in $A$ such that $p \leq f_2(p)$.
Then $f_1$ and $f_2$ are computable functions.
\end{Lemma}

{\bf Proof.}
$f_1$ and $f_2$ are computed by the algorithm below:
\begin{enumerate}
\item
Set $x$ equal to an arbitrary element of $A$ smaller than $p$.
\item
\label{Iteration}
\begin{enumerate}
\item
If $x=p$, output:
$f_1(p)=f_2(p)=x$.
Stop.
\item
If $x$ is a successor element and $p_S(x)\geq p$,
then output: $f_1(p)=x$ and $f_2(p)=p_S(x)$.
Stop.
\item
If $x$ is a successor element and $p_S(x)\leq p$,
set $x=p_S(x)$.
\item
If $x$ is a limit element and $x\neq p$, 
take the sequence
\[ \varphi_{q_S(x)}(0), \varphi_{q_S(x)}(1), \ldots. \]
Search for the smallest $i$ satisfying $\varphi_{q_S(x)}(i)\leq p$ and
set $x=\varphi_{q_S(x)}(i)$.
(Such $i$ exists because this sequence
is monotonically decreasing and converges to $x$ and $x<p$.)
\end{enumerate}
\item
Repeat step \ref{Iteration}.
\end{enumerate}

While this algorithm works, $x$ remains less
or equal to $p$.

From the definition of the system of notations it follows that 
the values of $f_1$ and $f_2$ output by the algorithm are correct.
It remains to prove that algorithm always outputs $f_1(p)$ and $f_2(p)$.

For a contradiction, assume that the algorithm
does not output $f_1(p)$ and $f_2(p)$ for some $p\in Q$.
This can happen only if the algorithm goes into eternal loop, i.e. if
Step \ref{Iteration} is executed infinitely many times.

Each execution of Step \ref{Iteration} increases the value of $x$.
Let $x_i$ be the value of $x$ after the $i^{\rm th}$
repetition of Step \ref{Iteration}.
Then $x_1, x_2, x_3, \ldots$ is an infinite 
monotonically increasing sequence.
This contradicts the set $A$ being well-ordered in decreasing order.
$\Box$

\subsection{Three examples}

\begin{figure*}
\label{Figure}
\begin{center}
\begin{picture}(380,250)
\multiput(20,20)(0,70){3}{\line(1,0){360}}
\multiput(20,18)(0,70){3}{\line(0,1){4}}
\multiput(380,18)(0,70){3}{\line(0,1){4}}
\multiput(20,35)(0,70){3}{\makebox(0,0){$0$}}
\multiput(380,35)(0,70){3}{\makebox(0,0){$1$}}
\multiput(20,18)(0,70){2}{\line(0,1){4}}
\multiput(270,35)(0,70){2}{\makebox(0,0){$\frac{2}{3}$}}
\multiput(270,18)(0,70){2}{\line(0,1){4}}
\multiput(246,35)(0,70){2}{\makebox(0,0){$\frac{3}{5}$}}
\multiput(246,18)(0,70){2}{\line(0,1){4}}
\multiput(234,18)(0,70){2}{\line(0,1){4}}
\multiput(229,18)(0,70){2}{\line(0,1){4}}
\multiput(226,18)(0,70){2}{\line(0,1){4}}
\multiput(223,18)(0,70){2}{\line(0,1){4}}
\multiput(220,18)(0,70){2}{\line(0,1){4}}
\multiput(218,18)(0,70){2}{\line(0,1){4}}
\multiput(216,18)(0,70){2}{\line(0,1){4}}
\multiput(214,18)(0,70){2}{\line(0,1){4}}
\multiput(212,18)(0,70){2}{\line(0,1){4}}
\multiput(210,18)(0,70){3}{\line(0,1){4}}
\multiput(210,35)(0,70){3}{\makebox(0,0){$\frac{1}{2}$}}
\multiput(225,35)(0,70){2}{\makebox(0,0){$\ldots$}}
\put(200,88){\line(0,1){4}}
\put(200,105){\makebox(0,0){$\frac{24}{49}$}}
\put(196,88){\line(0,1){4}}
\put(192,88){\line(0,1){4}}
\put(188,88){\line(0,1){4}}
\put(184,88){\line(0,1){4}}
\put(182,88){\line(0,1){4}}
\put(180,88){\line(0,1){4}}
\put(178,88){\line(0,1){4}}
\put(176,88){\line(0,1){4}}
\put(175,88){\line(0,1){4}}
\put(175,105){\makebox(0,0){$\frac{12}{25}$}}
\put(188,105){\makebox(0,0){$\ldots$}}
\put(100,105){\makebox(0,0){?}}
\put(175,18){\line(0,1){4}}
\put(168,18){\line(0,1){4}}
\put(163,18){\line(0,1){4}}
\put(159,18){\line(0,1){4}}
\put(157,18){\line(0,1){4}}
\put(155,18){\line(0,1){4}}
\put(153,18){\line(0,1){4}}
\put(151,18){\line(0,1){4}}
\put(150,18){\line(0,1){4}}
\put(150,35){\makebox(0,0){$\frac{4}{9}$}}
\put(165,35){\makebox(0,0){$\ldots$}}
\put(140,18){\line(0,1){4}}
\put(135,18){\line(0,1){4}}
\put(132,18){\line(0,1){4}}
\put(130,18){\line(0,1){4}}
\put(128,18){\line(0,1){4}}
\put(126,18){\line(0,1){4}}
\put(125,18){\line(0,1){4}}
\put(125,35){\makebox(0,0){$\frac{3}{7}$}}
\put(138,35){\makebox(0,0){$\ldots$}}
\put(75,35){\makebox(0,0){?}}
\put(100,158){\line(0,1){4}}
\put(100,175){\makebox(0,0){$\frac{1}{3}$}}
\put(80,158){\line(0,1){4}}
\put(80,175){\makebox(0,0){$\frac{1}{4}$}}
\put(68,158){\line(0,1){4}}
\put(62,158){\line(0,1){4}}
\multiput(57,158)(-4,0){3}{\line(0,1){4}}
\multiput(46,158)(-3,0){3}{\line(0,1){4}}
\multiput(37,158)(-2,0){5}{\line(0,1){4}}
\multiput(35,158)(-1,0){15}{\line(0,1){4}}
\put(35,175){\makebox(0,0){\ldots}}
\put(50,65){\makebox(0,0){\huge $\PFIN$}}
\put(42,135){\makebox(0,0){\huge $\FIN$}}
\put(37,205){\makebox(0,0){\huge $\EX$}}
\end{picture}
\end{center}
\caption{The probability hierarchies for $\EX$, $\FIN$ and $\PFIN$}
\end{figure*}
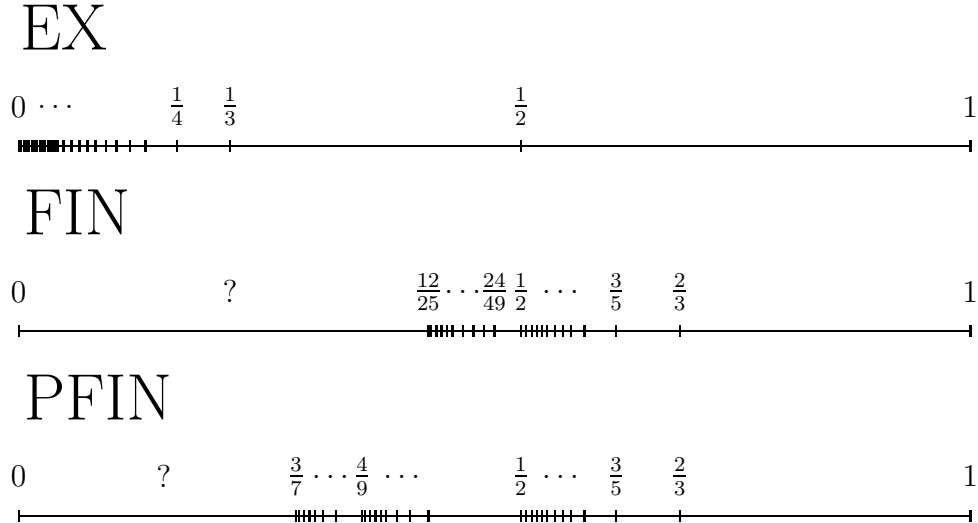


In Figure 1, 
we show the known parts of probability hierarchies
for three learning criteria:
\begin{itemize}
\item
$\EX$ (learning in the limit, Pitt and Smith\cite{Pitt,PS}),
\item
$\FIN$ (Freivalds\cite{Freivalds},
Daley, Kalyanasunda\-ram and Velauthapillai\cite{DKV92}), and 
\item
$\PFIN$ (Daley, Kalyanasundaram and Velauthapil\-lai\cite{DKV92a,DK93}).
\end{itemize}

We see that these probability hierarchies contain infinite decreasing
sequences but none of them contains an infinite increasing sequence.
Known parts of these hierarchies are well-ordered in decreasing
ordering.

We will show that, for $\PFIN$-type learning, the entire
hierarchy is well-ordered and will use this property to study
its properties.

\section{Decidability result}
\label{sec:decidable}

The outline for this section is as follows.
We start with describing a set $\AAA$
in two equivalent forms in subsection \ref{Subsection0}.
Then, in sections \ref{sub:technical} and \ref{sec:treeprop}, 
we show several technical lemmas about the set $\AAA$,
including the equivalence of the two descriptions.
Then, we show that $\AAA$ is the probability hierarchy for $\PFIN$.
The proof of that consists of two parts: diagonalization and simulation.
The diagonalization part is shown in subsection \ref{Subsection1}.
The simulation argument is more complicated.
First, in section \ref{Subsection2}, we show that $\AAA$ is well-ordered and
has a system of notations.
Finally, in subsection \ref{Subsection3}
we use these technical results to construct a universal simulation
argument.
Our diagonalization theorem uses methods from Kummer's
paper on $\PFIN$-teams\cite{Kummer94}
but the simulation part uses new techniques and
is far more complicated.

\subsection{Description of probability hierarchy}
\label{Subsection0}

Our description has two equivalent forms.
First, we describe it as a set of solutions to a
particular optimization problem on trees.

Similarly to \cite{Kummer94},
we define trees as finite nonempty subsets
of $\bbbn^{*}$ which are closed under initial
segments. The root of each tree 
is the empty string $\epsilon$.
A vertex $u$ is a child of a vertex $v$ if
$u=vn$ for some $n\in\bbbn$.
Next, we define labelings of trees by positive reals.
The definition below is equivalent to one in \cite{Kummer94}, 
with some minor technical modifications.

\begin{Definition}
\label{def:label}
Let $0<p<q$.
An $(p, q)$-labeling of a tree $T$ is a pair of mappings
$\nu_1, \nu_2 : T\rightarrow \bbbr^{+}$ such that
\begin{enumerate}
\item
\label{ite:label1}
$\nu_1(\epsilon)\geq p$ and $\nu_2(\epsilon)=0$,
\item
\label{ite:label2}
If $t_1, \ldots, t_s$ are all direct successors of $t$, 
then\footnote{Definition
in \cite{Kummer94} incorrectly uses $\nu_1(t)$ instead of
$\nu_1(t)+\nu_2(t)$ here.}
$\sum_{i=1}^s \nu_2(t_i)\leq \nu_1(t)+\nu_2(t)$
and $\nu_1(t_i)+\nu_2(t_i)\geq p$ for
$i=1, \ldots, s$,
\item
\label{ite:label3}
For each branch the sum of the $\nu_1$-labels of all of its nodes
is at most $q$.
\end{enumerate}
\end{Definition}


Labelings by natural numbers have an intuitive meaning.
$\nu_1(v)+\nu_2(v)$ is the number of machines that have
issued a conjecture consistent with the initial segment $v$.
In particular, $\nu_2(v)$ is the number of machines that
have issued such a conjecture on some prefix of $v$ and
$\nu_1(v)$ is the number of machines that have output
it after seeing the whole segment $v$.

Then, the requirements of definition have the following interpretation.
$\nu_1(t)+\nu_2(t)\geq p$ means that, for every segment $t$ in the
tree, there must be at least $p$ machines with conjectures
consistent with $t$. 

The second requirement, $\sum_{i=1}^s \nu_2(t_i)\leq \nu_1(t)+\nu_2(t)$ 
has the following interpretation. $\nu_2(t_i)$ is the number
of machines which have issued a conjecture consistent
with $t_i$ after reading a prefix of $t_i$. 
A conjecture consistent with $t_i$ is also consistent with $t$.
A prefix of $t_i$ could be either $t$ or a prefix of $t$. 
Since a conjecture can be only consistent with
one of segments $t_i$, $\sum_{i=1}^s \nu_2(t_i)$ must be at
most the total number of machines which have issued a conjecture
consistent with $t$ after reading either $t$ or a prefix of $t$.
The number of such machines is $\nu_1(t)+\nu_2(t)$.

\begin{figure}
\label{fig-tree}
\begin{center}
\epsfxsize=4in
\hspace{0in}
\epsfbox{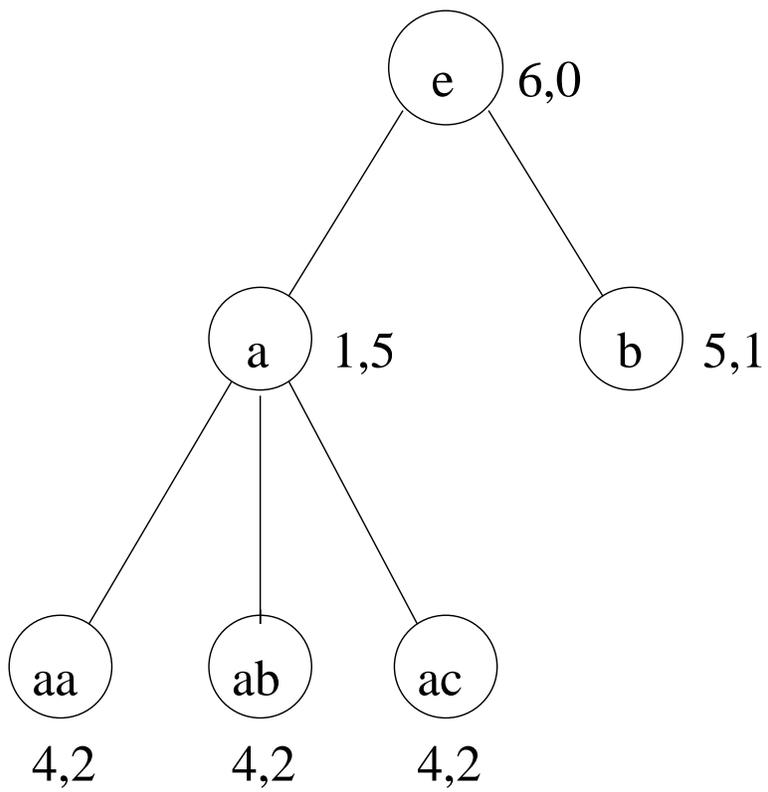}
\caption{\it A tree and a $(6, 11)$-labeling for it.}
\end{center}
\end{figure}

Finally, the third requirement means that the total number of machines that
issue conjectures on any branch is at most $q$.
An example of a labeling is shown in Figure \ref{fig-tree}.
The first number near node is $\nu_1(t)$, the second number 
is $\nu_2(t)$.

Labelings with reals have a similar interpretation, with
$\nu_1(t)$ and $\nu_2(t)$ being the probabilities that
a probabilistic machine has output a conjecture consistent with $t$.


Let $p_T$ denotes the largest number $p$ such that
there is a $(p, 1)$ labeling of $T$. 
(For the tree in Figure \ref{fig-tree}, $p=6/11$.)
Let
\[ \AAA = \{p_T | T \mbox{ is a tree}\}. \]

The second description is algebraic, by a recurrence relation.
Let set $\AAB$ defined by the following rules:
\begin{enumerate}
\item
$1\in \AAB$;
\item
\label{Rule}
If $p_1, p_2, \ldots, p_s \in \AAB$ and $p\in[0, 1]$ is a number such that
there exist $q_1, \ldots, q_s\in[0,1]$ satisfying
\begin{enumerate}
\item
$q_1+q_2+\ldots+q_s=p$;
\item
$\frac{p}{q_i+1-p}=p_i$ for $i=1, \ldots, s$,
\end{enumerate}
then $p\in \AAB$;
\end{enumerate}

In section \ref{sec:treeprop}, we will show that both definitions
give the same set $\AAA=\AAB$. After that, we will prove
that $\AAA$ is the probability hierarchy for $\PFIN$.

\subsection{Technical lemmas: algebraic description}
\label{sub:technical}

In this subsection, we study the properties of the rule that
generates the set $\AAB$.
The results of this subsection are used in
various parts of section \ref{sec:decidable}.
First, we show that the rule \ref{Rule} can be
described without using variables $q_i$.


\begin{Lemma}
\label{Formula}
If there exist $q_1, \ldots, q_s\in[0, 1]$ satisfying
$q_1+q_2+\ldots+q_s=p$ and
$\frac{p}{q_i+1-p}=p_i$ for $i=1, \ldots, s$, then
\begin{equation}
\label{eq:rule}
p=\frac{s}{(s-1)+\sum_{i=1}^s{\frac{1}{p_i}}}.
\end{equation}
\end{Lemma}

\noindent
{\bf Proof.}
$\frac{p}{q_i+1-p}=p_{i}$ is equivalent to $q_i=\frac{p}{p_{i}}+p-1$.
Hence,
\[ p=\sum_{i=1}^s q_i=\sum_{i=1}^s(\frac{p}{p_i}+p-1)=
\left(\sum_{i=1}^s\frac{1}{p_{i}}\right)p+s\cdot p-s ,\]
\[ s= \left(\sum_{i=1}^s\frac{1}{p_{i}}\right)p+(s-1)p ,\]
\[ p=\frac{s}{(s-1)+\sum_{i=1}^s(\frac{1}{p_{i}})} .\]
$\Box$

We shall use both forms of the rule \ref{Rule}.
The rule with $q_i$ is more natural in simulation and diagonalization
arguments but is less convenient for algebraic manipulations.
We also use a version of Lemma \ref{Formula1} where equality
is replaced by inequality.

\begin{Lemma}
\label{Formula1}
If there exist $q_1, \ldots, q_s\in[0, 1]$ satisfying
$q_1+q_2+\ldots+q_s=p$ and
$\frac{p}{q_i+1-p}\leq p_i$ for $i=1, \ldots, s$, then
\begin{equation}
\label{eq:rule1}
p\leq\frac{s}{(s-1)+\sum_{i=1}^s{\frac{1}{p_i}}}.
\end{equation}
\end{Lemma}

\noindent
{\bf Proof.}
Similar to the proof of Lemma \ref{Formula}, with
$\leq$ or $\geq$ instead of $=$ where necessary.
$\Box$

Lemma \ref{Formula} suggests that the rule \ref{Rule}
can be considered as a function of $p_1, \ldots, p_s$.
Next lemmas show that this function is monotonic
and continuous.

\begin{Lemma}
\label{Monotonicity}
If 
\begin{enumerate}
\item
$p\in \AAB$ follows from $p_1\in \AAB, \ldots, 
p_s\in \AAB$ by rule \ref{Rule};
\item
$p'\in \AAB$ follows from $p'_1\in \AAB, \ldots, 
p'_s\in \AAB$ by rule \ref{Rule};
\item
$p_1\leq p'_1, \ldots, p_s\leq p'_s$,
\end{enumerate}
then $p\leq p'$.
If $p_i<p'_i$ for at least one $i$, then $p<p'$.
\end{Lemma}

\noindent
{\bf Proof.}                   
By Lemma \ref{Formula}
\[ p=\frac{s}{(s-1)+\sum_{i=1}^s{\frac{1}{p_i}}} \mbox{ and }
 p'=\frac{s}{(s-1)+\sum_{i=1}^s{\frac{1}{p'_i}}} .\]
From $p_i\leq p'_i$ it follows that
$\frac{1}{p_i}\geq \frac{1}{p'_i}$ and
\[ (s-1)+\sum_{i=1}^s{\frac{1}{p_i}}\geq 
 (s-1)+\sum_{i=1}^s{\frac{1}{p'_i}} ,\]
\[ p=\frac{s}{(s-1)+\sum_{i=1}^s{\frac{1}{p_i}}}\leq 
 \frac{s}{(s-1)+\sum_{i=1}^s{\frac{1}{p'_i}}}=p' .\]
If $p_i<p'_i$ for some $i$,
then $1/p_i>1/p'_i$ and all inequalities are strict.
$\Box$

\begin{Lemma}
\label{Limits}
Let $p_j=\lim_{i\rightarrow\infty} p_{j, i}$
and $r=\lim_{i\rightarrow\infty} r_i$.
If, for all $i\in\bbbn$, 
$r_i\in \AAB$ follows from $p_{1, i}\in \AAB, \ldots, p_{s, i}\in \AAB$
by rule \ref{Rule}, then
$r\in \AAB$ follows from $p_1\in \AAB, \ldots, p_s\in \AAB$ 
by rule \ref{Rule}.
\end{Lemma}

\noindent
{\bf Proof.}

\[ r=\lim_{i\rightarrow\infty} r_i=
\lim_{i\rightarrow\infty} \frac{s}{(s-1)+\sum_{j=1}^s{\frac{1}{p_{j, i}}}} =\]
\[ \frac{s}{(s-1)+\sum_{j=1}^s{\frac{1}{\lim_{j\rightarrow\infty} p_{j, i}}}} =
\frac{s}{(s-1)+\sum_{j=1}^s{\frac{1}{p_j}}} .\]
$\Box$

The last result of this section
relates the numbers generated by
applications of the rule \ref{Rule} to
$p_1\in \AAB$, $\ldots$, $p_s\in \AAB$ and
$\frac{p_1}{1+p_1}\in \AAB$, $\ldots$, $\frac{p_s}{1+p_s}\in \AAB$.

\begin{Lemma}
\label{lem:rule}
An application of the rule \ref{Rule} to
$x_1\in \AAB$, $\ldots$, $x_s\in \AAB$ generates $p\in \AAB$ if and only if
an application of the rule \ref{Rule} to
$\frac{x_1}{1+x_1}\in \AAB$, $\ldots$, $\frac{x_s}{1+x_s}\in \AAB$ 
generates $\frac{p}{1+p}\in \AAB$.
\end{Lemma}

\noindent
{\bf Proof.}
Assume that equation (\ref{eq:rule}) 
is true for $p_1=x_1$, $\ldots$, $p_s=x_s$.
Then,
\[ \frac{p}{1+p}=
\frac{\frac{s}{(s-1)+\sum_{i=1}^s \frac{1}{x_i}}}
{1+\frac{s}{(s-1)+\sum_{i=1}^s \frac{1}{x_i}}}=
\frac{s}{(s-1)+\sum_{i=1}^s \frac{1}{x_i}+s}=\]
\[\frac{s}{(s-1)+\sum_{i=1}^s (1+\frac{1}{x_i})}=
\frac{s}{(s-1)+\sum_{i=1}^s \frac{1+x_i}{x_i}} .\]
This is precisely equation (\ref{eq:rule}) 
for $p_1=\frac{x_1}{1+x_1}$, 
$\ldots$, $p_s=\frac{x_s}{1+x_s}$.

The opposite direction (equation (\ref{eq:rule}) 
for $p_1=\frac{x_1}{1+x_1}$, 
$\ldots$, $p_s=\frac{x_s}{1+x_s}$ implies
equation (\ref{eq:rule}) is true for $p_1=x_1$, $\ldots$, $p_s=x_s$)
is similar.
$\Box$

\subsection{Technical lemmas: tree description}
\label{sec:treeprop}

We start by showing that for a tree $T$ and its subtrees $T_i$,
$p_T$ and $p_{T_i}$ are related similarly to rule \ref{Rule}.

\begin{Lemma}
\label{lem:mult}
Let $r>0$ and $T$ be a tree with $(p, q)$-labeling. 
Then, there is a $(pr, qr)$-labeling for $T$.
\end{Lemma}

\noindent
{\bf Proof.}
We multiply all labels 
by $r$ and obtain a $(pr, qr)$-labeling.
$\Box$

\begin{Lemma}
\label{lem:existlabel}
Let $t_1, \ldots, t_s$ be all direct successors of the root
in a tree $T$ and $T_1, T_2, \ldots, T_S$ be the subtrees with
roots $t_1, t_2, \ldots, t_s$.
Assume there are $q_1$, $\ldots$, $q_s$ such that $\sum_{i=1}^s q_i=p$
and 
\[ p=p_{T_i}(q_i+1-p) \]
for $i\in\{1, \ldots, s\}$.
Then $p_T=p$.
\end{Lemma}

\noindent
{\bf Proof.}
First, we construct a $(p, 1)$-labeling.
Let $\nu^i_1$, $\nu^i_2$ be a $(p_{T_i}, 1)$-labeling 
for $T_i$. 
We define
\[ \nu_1(t)=\cases{p, &if $t=\epsilon$\cr
		p-q_i, &if $t=t_i$\cr
		(1+q_i-p)\nu^i_1(t), &if $t$ is a descendant of $t_i$\cr} \]
\[ \nu_2(t)=\cases{0, &if $t=\epsilon$\cr
		q_i, &if $t=t_i$\cr
		(1+q_i-p)\nu^i_2(t), &if $t$ is a descendant of $t_i$\cr} \]
Properties \ref{ite:label1} and \ref{ite:label2}
can be checked directly from the definitions of $\nu_1$ and $\nu_2$.

We prove Property \ref{ite:label3}.
Let $u$ be a direct successor of $t_i$.
Then, the sum of $\nu^i_1$-labels on any branch
starting at $u$ is at most $1-p_{T_i}$.
(By Property \ref{ite:label3} of $\nu^i_1$, it is at most $1$
for any branch starting at $t_i$ and $\nu^i_1(t_i)\geq p_{T_i}$.)
Hence, the sum of $\nu$-labels for such a branch
is at most $(q_i+1-p)(1-p_{T_i})$.
A branch starting at $\epsilon$ consists of $\epsilon$,
$t_i$ and a branch starting at a direct descendant of $t_i$.
Hence, the sum of all its $\nu_1$-labels is at most
\[ p+(p-q_i)+(q_i+1-p)(1-p_{T_i})=p+1-(1+q_i-p)+(q_i+1-p)(1-p_{T_i})=\]
\[ p+1-(q_i+1-p)p_{T_i} = p+1-p =1 .\]

For a contradiction, assume that 
there is $p'>p$ and a $(p', 1)$-labeling $(\nu'_1, \nu'_2)$ for $T$.
Let $q'_i=\nu'_2(t_i)$.
If we restrict ourselves to the subtree $T_i$ and 
add $\nu'_2(t_i)$ to $\nu'_1(t_i)$,
we obtain a $(p', 1-p'+q'_i)$-labeling for $T_i$.
By Lemma \ref{lem:mult}, there is 
a $(p'/(1-p'+q'_i), 1)$ labeling for $T_i$.
Hence, 
\[ \frac{p'}{1-p'+q'_i}\leq p_{T_i} =\frac{p}{1-p+q_i}<\frac{p'}{1-p+q_i} ,\]
\[ (1-p'+q'_i)>(1-p+q_i), \]
\[ p'-q'_i<p-q_i .\]
We consider the sum of these expressions for all $i$.
\[ (s-1)p' \leq s\cdot p'-\sum_{i=1}^s q'_i = \sum_{i=1}^s (p'-q'_i)
 < \sum_{i=1}^s (p-q_i) =\]
\[= s\cdot p-\sum_{i=1}^s q_i= (s-1)p \]
and $p'< p$.
Contradiction, proving the lemma.
$\Box$

By Lemma \ref{Formula}, the relation between
$p_T$ and $p_{T_1}$, $\ldots$, $p_{T_s}$ is also expressed
by the equation (\ref{eq:rule}).
We can now show the equivalence of the two definitions.

\begin{Lemma}
\label{lem:treeequiv}
$\AAA=\AAB$.
\end{Lemma}

\noindent
{\bf Proof.}
By induction.
If $p\in \AAB$ follows from $p_1, \ldots, p_s\in \AAB$ by rule \ref{Rule}
and $p_{T_i}=p_i$ for trees $T_i$,
we construct a tree $T$ consisting of the root, $T_1, \ldots, T_s$
and make the roots of $T_1$, $T_2$, $\ldots$, $T_s$ 
children of $T$'s root.
Then, $p_T=p$ (by Lemma \ref{lem:existlabel}).
Hence, for any $p\in \AAB$, there is a tree $T$ with $p_T=p$ .
This means $\AAB\subseteq \AAA$.

Similarly, we can show that $p_T\in \AAB$ for any tree $T$.
$\Box$

Next, we show that the $(p_T, 1)$-labeling
of Lemma \ref{lem:existlabel} uses only rational numbers and,
hence, can be transformed into a labeling that uses only integers.

\begin{Lemma}
\label{lem:rational1}
For any tree $T$, $p_T\in\bbbq$.
\end{Lemma}

\noindent
{\bf Proof.}
By induction over the depth of $T$.
For a tree consisting of root only, $p=1$.

Otherwise, let $t_1, \ldots, t_s$ be 
all direct successors of the root in $T$
and $T_1, T_2, \ldots, T_S$ be the subtrees with
roots $t_1, t_2, \ldots, t_s$.
The depth of these subtrees is smaller than
the depth of $T$. Hence, all $p_{T_i}$ are rationals.
Equation (\ref{eq:rule}) implies that $p_T$ is rational, too.
$\Box$

\begin{Lemma}
\label{lem:rational2}
$(p_T, 1)$-labeling constructed in the proof of Lemma \ref{lem:existlabel}
uses only rational numbers.
\end{Lemma}

\noindent
{\bf Proof.}
By induction over the depth of $T$.
Again, the lemma is evident for the tree with the root only.

For other trees, notice that
all $q_i$ can expressed by $p$ and $p_{T_i}$.
Hence, $q_1$, $\ldots$, $q_s$ are rationals.
Label of the root is the rational number $p$,
labels of $t_1, \ldots, t_s$ are rationals $p-q_1, \ldots, p-q_s$
and labels of other nodes are $(1-p+q_i)\nu^i_j(t)$.
$(1-p+q_i)$ is a rational number because $p$ and $q_i$ are rationals
and $\nu^i_j(t)$ is a rational number because 
$\nu^i_j$ is a part of the $(p_{T_i}, 1)$-labeling
for a tree of smaller depth.
$\Box$

\begin{Corollary}
\label{lem:natural}
Let $T$ be a tree.
Then there is $n\in\bbbn$ such that $T$ has 
$(p_T n, n)$-labeling with labels from $\bbbn$.
\end{Corollary}

\noindent
{\bf Proof.}
Let $n$ be the least common denominator of all rational
numbers in the $(p_T, 1)$-labeling $\nu_1$,
$\nu_2$ of Lemma \ref{lem:existlabel}.
Then, $n\nu_1(t), n\nu_2(t)$ is a $(p_T n, n)$ labeling and
uses only natural numbers.
$\Box$

\subsection{Universal diagonalization}
\label{Subsection1}

Let $0^j$ denote a sequence of $j$ zeros and
$0^{\omega}$ denote an infinite sequence of zeros.
Let $K$ be the halting set, i.e. the set of all $i$ such that
program $\varphi_i$ halts on input $i$. 
Let $K_s$ be the set of all $i$ such that $\varphi_i$ halts
on input $i$ in at most $s$ steps. 
For a set $S$, let $\chi_S$ be the characteristic function of $S$:
$\chi_S(i)=1$ if $i\in S$ and $\chi_S(i)=0$ otherwise.
					
\begin{Definition}
\cite{Kummer94}
Let $T$ be a tree of depth $d$.
$S_T$ is the set of all recursive functions $f$ such that the sequence of
values $f(0)$, $f(1)$, $\ldots$ is of the form
\[ i_1\ldots i_d 0^{t_1}a_1 0^{t_2} a_2 \ldots 0^{t_l}a_l 0^{\omega} \]
where each $t_h=\min \{t : |\{j : i_j\in K_t\}|\geq h\}$
is finite, $(a_1, \ldots, a_l)\in T$, and either $l=|\{ j: i_j\in K\}|$
or $(a_1, \ldots, a_l)$ is a leaf of $T$.
\end{Definition}

\begin{Lemma}
\label{lem:ident}
\cite{Kummer94}
If $T$ has an $(m, n)$-labeling by integers then
\[S_T\in [m, n]\PFIN.\]
\end{Lemma}

The next lemma is an extension of Kummer's results to probabilistic learning.
The proof is similar to Theorem 16 in \cite{Kummer94}.
We give it here for completeness.

\begin{Lemma}
\label{lem:nonidentA}
If $S_T\in \langle p\rangle\PFIN[O]$ and $K$ is not
Turing reducible to $O$, then $T$ has a $(p-\epsilon, 1)$ labeling
for any $\epsilon>0$.
\end{Lemma}

\noindent
{\bf Proof.}
Let $k$ be the depth of $T$.
Let $M$ denote an IIM that identifies $S_T$ with the $O$-oracle.
For arbitrary $i_1, \ldots, i_k$, we enumerate a set $T_{i_1, \ldots, i_k}$.

Define the event $P(c, s)$ to be true 
iff  $c=|\{j : i_j \in K_s\}|$ and, for each
$(a_1, \ldots, a_c)\in T$ and
$\sigma_c=i_1\ldots i_k 0^{t_1}a_1 0^{t_2} a_2 \ldots 0^{t_c}a_c$,
the probability that $M^O$ outputs a program computing a function
with an initial segment $\sigma_c$ while reading $\sigma_c 0^s$
is at least $p-\epsilon$.

The procedure for enumerating $T_{i_1, \ldots, i_k}$ is as follows. 

{\em Initialization.}
Let $t=0, c'=-1, T_{i_1, \ldots, i_k}=\emptyset$.

{\em Step l.}
Search for the smallest $s>t$ satisfying $P(c, s)$ for some $c>c'$.
If the search terminates, enumerate 
$(\chi_{K_s}(i_1), \ldots, \chi_{K_s}(i_k))$ into $T_{i_1, \ldots, i_k}$,
set $t=s$, $c'=c$ and go to Step $l+1$.

\begin{Claim}
$(\chi_K(i_1), \ldots, \chi_K(i_k))\in T_{i_1, \ldots, i_k}$.
\end{Claim}

\noindent
{\bf Proof.}
Let $c=|\{j : i_j \in K\}|$.
$P(c, s)$ holds for all sufficiently large $s$ because $M^O$ infers
all functions $\sigma_c 0^{\infty}$.
After discovering it, $(\chi_K(i_1), \ldots, \chi_K(i_k))=
(\chi_{K_s}(i_1), \ldots, \chi_{K_s}(i_k))$ is enumerated into
$T_{i_1, \ldots, i_k}$.
$\Box$

\begin{Claim}
$|T_{i_1, \ldots, i_k}|=k+1$ for some $i_1, \ldots, i_k$.
\end{Claim}

\noindent
{\bf Proof.}
If $(\chi_K(i_1), \ldots, \chi_K(i_k))\in T_{i_1, \ldots, i_k}$
and $|T_{i_1, \ldots, i_k}|\leq k$ for all $i_1, \ldots, i_k$, 
then, by Fact 6 in \cite{Kummer94}, $K$ is Turing-reducible to $O$.
$\Box$

Hence, there exists $i_1, \ldots, i_k$ and  
$s_1 < \ldots < s_{k+1}$ such that $P(l-1, s_l)$
for $l=1, \ldots, k+1$.
Define the label $\nu_1(\tau)$ of $\tau=(a_1, \ldots, a_{l-1})$
as the probability that:
\begin{enumerate}
\item
$M$ does not output a program while reading $\sigma_{l-2}0^{s_{l-2}}$, 
where $\sigma_c=i_1\ldots i_k 0^{t_1}a_1 0^{t_2} a_2 \ldots 0^{t_c}a_c$,
and
\item
$M$ outputs a program computing a function with 
the initial segment $\sigma_{l-1}$ while reading 
$\sigma_{l-1}0^{s_{l-1}}$.
\end{enumerate}
For $\tau=\epsilon$, there is no segment $\sigma_{-1}$ and
$\nu_1(\epsilon)$ is just the probability that 
$M^O$ outputs a program computing a function with 
the initial segment $\sigma_0$ while reading 
$\sigma_0 0^{s_0}$.

The label $\nu_2(\tau)$ is $0$ for $\tau=\epsilon$
and the probability that $M^O$ outputs a program
computing a function with
the initial segment $\sigma_{l-1}$ while reading 
$\sigma_{l-2}0^{s_{l-2}}$ for $\tau=(a_1, \ldots, a_{l-1})$.

Next, we verify that all conditions of Definition \ref{def:label}
are satisfied.
Property \ref{ite:label1} follows from the definitions of
$\nu_1(\epsilon)$, $\nu_2(\epsilon)$ and $P(0, s)$.

For property \ref{ite:label2},
notice that $\nu_1(t)+\nu_2(t)$ is the total probability
that $M^O$ outputs a function consistent with $\sigma_{l-1}$
while reading $\sigma_{l-1}0^{s_{l-1}}$.
$\nu_1(t_i)$ are the probabilities that a particular
continuation of $\sigma_{l-1}$ is an initial segment of the function.
These events are mutually exclusive.
Hence, $\sum_{i=1}^s \nu_1(t_i) \leq \nu_1(t)+\nu_2(t)$.
$\nu_1(t_i)+\nu_2(t_i)\geq p-\epsilon$ is true because
$M^O$ outputs a program consistent with $\sigma_l 0^{s_l}$
with a probability at least $p-\epsilon$ (by the definition of $P(c, s)$).

Property \ref{ite:label3} is true because the sum of all $\nu_1$-labels on
any branch is at most the probability that $M^O$ outputs a conjecture
while reading $\sigma_k 0^{s_k}$ and, hence, is at most 1.
$\Box$

If there is no oracle $O$, we get

\begin{Corollary}
\label{lem:nonident}
If $S_T\in \langle p\rangle\PFIN$, then $T$ has a $(p-\epsilon, 1)$ labeling
for any $\epsilon>0$.
\end{Corollary}

\begin{Corollary}
\label{lem:treesets}
For a tree $T$, $S_T\in\langle p_T\rangle\PFIN$ and
$S_T\notin\langle p_T+\epsilon\rangle\PFIN$
for any $\epsilon>0$.
\end{Corollary}

\noindent
{\bf Proof.}
Corollary \ref{lem:natural} and Lemma \ref{lem:ident} imply that $S_T\in[p_T n, n]\PFIN$ for
appropriate $n$. A $[p_T n, n]\PFIN$ team can be simulated by a $\langle p_T\rangle\PFIN$
probabilistic machine that chooses one of $n$ machines in the team equiprobably.

If $S_T\in \langle p_T+\epsilon\rangle\PFIN$, then, 
there is a $(p_T+\epsilon/2, 1)$ labeling of $T$
(Corollary \ref{lem:nonident}).
This is impossible because $p_T$ is the largest number
such that there is a $(p_T, 1)$ labeling of $T$.
$\Box$

\begin{Theorem}
\label{Part1}
If $p, q\in \AAA$ and $p\neq q$, then $\langle p\rangle\PFIN\neq\langle q\rangle\PFIN$.
\end{Theorem}

\noindent
{\bf Proof.}
Follows from Corollary \ref{lem:treesets} and Lemma \ref{lem:treeequiv}.
$\Box$

\subsection{Well-ordering and system of notations}
\label{Subsection2}

It remains to prove that, for any probability $p$, 
$\PFIN\langle p\rangle$-type learning is equivalent to $PFIN$-type
learning with some probability belonging to $\AAA$.
Our diagonalization technique was similar to \cite{Kummer94}.
The simulation part is more complicated.
Simulation techniques in \cite{Kummer94}
rely on fact that each team issues finitely many conjectures
and, hence, there are finitely many possible 
behaviors of these conjectures.
A probabilistic machine can issue infinitely many conjectures
and these conjectures have infinitely many possible behaviors.
This makes simulation far more complicated.

We need an algorithmic structure for manipulating 
an infinite number of possibilities.
We establish it by proving that $\AAA$ is well-ordered and has 
a system of notations.

\begin{Theorem}
\label{Basic}
The set $\AAA$ is well-ordered in decreasing ordering and has a system 
of notations.
\end{Theorem}

\noindent
{\bf Proof.}
We construct a system of notations for the set $\AAA$ inductively.
First, we construct a system of notations for $\AAA\cap [\frac{1}{2},1]$.
Then we extend it, obtaining system of notations
for $\AAA\cap [\frac{1}{3},1]$, $\AAA\cap [\frac{1}{4},1]$ and so on.

Freivalds\cite{Freivalds} proved
\[ \AAA\cap \left[\frac{1}{2},1 \right] = 
   \left\{\frac{1}{2}\right\}\cup\left\{\frac{n}{2n-1} |
   n\in\bbbn \& n\geq 1\right\}.\]
A system of notations for $\AAA\cap [\frac{1}{2},1]$ 
can be easily constructed from this description.
Below, we show how to construct a system of notations for 
$\AAA\cap [\frac{1}{n+1},1]$ using a system of notations
for $\AAA\cap [\frac{1}{n},1]$.

An outline of our construction is as follows:
\begin{enumerate}
\item
Split the segment $[\frac{1}{n+1}, \frac{1}{n}]$ into
smaller segments $[r_{i+1}, r_i]$
so that, if $p\in [r_{i+1}, r_i]$ and
$p\in \AAA$ follows from the rule \ref{Rule}, then 
$p_1\geq r_i, \ldots, p_s\geq r_i$.
(This property allows us to obtain a system of notations for 
$\AAA\cap [r_{i+1}, r_i]$ from a given system of notations 
for $\AAA\cap [r_i, 1]$ without using any knowledge about 
$\AAA\cap [r_{i+1}, r_i]$.)
We give the splitting and prove its properties in subsection 
\ref{subsub:split}.
\item
Using transfinite induction over the segments $[r_{i+1}, r_i]$, extend
the system of notations for $\AAA\cap [\frac{1}{n},1]$
to larger and larger segments $\AAA\cap [r_{i+1}, 1]$,
finally obtaining a system of notations for $\AAA\cap [\frac{1}{n+1},1]$.
This part is described in subsections \ref{subsub:transfinite},
\ref{subsub:distinguishing}, \ref{subsub:xdminimal} and \ref{subsub:system}.
\end{enumerate}

\subsubsection{Splitting the segment $[\frac{1}{n+1}, \frac{1}{n}]$}
\label{subsub:split}

The splitting consists of two steps.
\begin{enumerate}
\item
First, we take 
$\frac{p}{1+p}$ for $p\in \AAA\cap [\frac{1}{n}, \frac{1}{n-1}]$.
By Lemma \ref{lem:rule}, all $\frac{p}{1+p}$ belong to the set $\AAA$.
These points split $[\frac{1}{n+1}, \frac{1}{n}]$
into segments $[\frac{p}{1+p}, \frac{r}{1+r}]$.
\item
Each segment $[\frac{p}{1+p}, \frac{r}{1+r}]$
is split further by the sequence
\[ r_0=\frac{r}{1+r}, \mbox{~~~~}, 
r_{i+1}=\frac{2}{1+\frac{1}{p}+\frac{1}{r_i}}. \]
$r_0, r_1, r_2, \ldots$
is a monotonically decreasing sequence converging to $\frac{p}{1+p}$.
It splits $[\frac{p}{1+p}, \frac{r}{1+r}]$
into segments $[r_1, r_0]$, $[r_2, r_1]$, $\ldots$.
\end{enumerate}
Let $A_n$ denote the set consisting of all $\frac{p}{1+p}$
and $r_0, r_1, \ldots$ for all segments $[\frac{p}{1+p}, \frac{r}{1+r}]$.
Next, we prove several properties of the segments $[r_{i+1}, r_i]$ that
will be used further.

\begin{Lemma}
\label{p1First}
Let $[\frac{p}{1+p}, \frac{r}{1+r}]$ be a segment obtained
in the first step of the splitting.
If $x\in \AAA$ follows from $p_1, \ldots, p_s\in \AAA$ by the rule \ref{Rule}
and $x\in[\frac{p}{1+p}, \frac{r}{1+r}]$ then
\[ p_1\leq p, p_2\leq p, \ldots, p_s\leq p.\]
\end{Lemma}

\noindent
{\bf Proof.}
We have
\[ p_j = \frac{x}{1-x+q_j} < \frac{x}{1-x+0} = \frac{x}{1-x}, \]
\[ p_j(1-x)< x,\]
\[ p_j< x(1+p_j), \]
\[ \frac{p_j}{1+p_j}< x.\]
Therefore,
$\frac{p_j}{1+p_j}\leq \frac{p}{1+p}$ and $p_j\leq p$.
$\Box$

\begin{Lemma}
\label{p1Second}
Let $x\in \AAA\cap [r_{i+1}, r_i]$.
If $x\in \AAA$ follows from $p_1, \ldots, p_s\in \AAA$ by the rule \ref{Rule},
then
\[ p_1\geq r_i, p_2\geq r_i, \ldots, p_s\geq r_i.\]
\end{Lemma}

\noindent
{\bf Proof.}
We prove $p_1\geq r_i$ only.
($p_2\geq r_i, \ldots$ are proved similarly.)

Assume that $[r_{i+1}, r_i]$ was obtained by 
splitting $[\frac{p}{1+p}, \frac{r}{1+r}]$.
Then, $p_1\leq p, p_2\leq p, \ldots, p_s\leq p$
(Lemma \ref{p1First}).

From 
\[ \frac{x}{1-x+q_j}=p_j \]
it follows that
\[ q_j=\frac{x}{p_j}-1+x .\]
We have $p_2\leq p$. Hence,
\[ q_2\geq \frac{x}{p}-1+x, \]
\[ q_1\leq x-q_2 \leq 1-\frac{x}{p}, \]
\[ p_1=\frac{x}{1-x+q_1}\geq \frac{x}{2-x-\frac{x}{p}} =
 \frac{1}{\frac{2}{x}-1-\frac{1}{p}} .\]
From $x\in [r_{i+1}, r_i]$ we have that $x\geq r_{i+1}$ and
\[ p_1 \geq \frac{1}{\frac{2}{x}-1-\frac{1}{p}}\geq 
 \frac{1}{\frac{2}{r_{i+1}}-1-\frac{1}{p}}\geq 
 \frac{1}{(1+\frac{1}{r_i}+\frac{1}{p})-1-\frac{1}{p}}=r_i .\]
$\Box$

We have proved that all $x\in \AAA\cap[r_{i+1}, r_i]$ are generated
by applications of the rule \ref{Rule}
to $p_1, \ldots, p_s\in \AAA\cap[r_i, 1]$.
The next lemma bounds the number $s$.

\begin{Lemma}
\label{p1Quantity}
Let $x\in \AAA\cap [r_{i+1}, r_i]$, with $[r_{i+1}, r_i]$
being a segment obtained by splitting $[\frac{p}{1+p}, \frac{r}{1+r}]$.
If $x\in \AAA$ follows from $p_1, \ldots, p_s\in \AAA$ by the rule \ref{Rule},
then
\[ s \leq \frac{x}{\frac{x}{p}+x-1} .\]
\end{Lemma}

\noindent
{\bf Proof.}
From Lemma \ref{p1First} we have 
\[ q_j= \frac{x}{p_j}+x-1 \geq \frac{x}{p}+x-1 .\]
Hence,
\[ x=\sum_{j=1}^{s}q_j\geq s \left( \frac{x}{p}+x-1 \right), \]
\[ s\leq  \frac{x}{\frac{x}{p}+x-1}  .\]
$\Box$

\subsubsection{Well-ordering}
\label{subsub:transfinite}

\begin{Lemma}
$A_n$ is well-ordered.
\end{Lemma}

\noindent
{\bf Proof.}
$\AAA\cap [\frac{1}{n},1]$ is well-ordered by inductive assumption.
Hence, $\AAA\cap [\frac{1}{n}, \frac{1}{n-1}]$ is well-ordered, too.
The set $\{ \frac{p}{1+p} | p\in \AAA\cap [\frac{1}{n}, \frac{1}{n-1}]\}$
is order-isomorphic to $\AAA\cap [\frac{1}{n}, \frac{1}{n-1}]$.
Hence, it is well-ordered and the set of segments
$[\frac{p}{1+p}, \frac{q}{1+q}]$ into which
it splits $[\frac{1}{n+1}, \frac{1}{n}]$ is well-ordered, too.

$A_n$ is obtained by replacing
each segment $[\frac{p}{1+p}, \frac{r}{1+r}]$ 
with the sequence $r_0, r_1, \ldots$.
Each sequence is well-ordered.
Hence, the entire set $A_n$ is well-ordered.
$\Box$

Hence, we can use transfinite induction over this set.

\begin{Lemma}
\label{InductionStep}
$\AAA\cap [\frac{1}{n+1}, \frac{1}{n}]$
is well-ordered in decreasing ordering.
\end{Lemma}

\noindent
{\bf Proof.}
By transfinite induction over $A_n$.

\noindent
{\em Base case.}
The set $\AAA\cap [\frac{1}{n}, 1]$ is well-ordered.

\noindent
{\em Inductive case.}
Let $x\in A_n$.
We assume that $\AAA \cap [x', 1]$ is well-ordered for all $x'>x$, 
$x'\in A_n$ and prove that $\AAA\cap [x, 1]$ is well-ordered, too.
There are three cases:
\begin{enumerate}
\item
$x=\frac{p}{1+p}$ for $p\in \AAA\cap[\frac{1}{n+1}, \frac{1}{n}]$
and $p$ is a limit element.

Let $p$ be the limit of $p_1, p_2, \ldots$.
Then, $\frac{p}{1+p}$ is the limit of
$\frac{p_1}{1+p_1}, \frac{p_2}{1+p_2}, \ldots$
because the function $\frac{x}{1+x}$ is continuous.
By inductive assumption, each $[\frac{p_i}{1+p_i}, 1]$ is well-ordered.
Hence, their union $[\frac{p}{1+p}, 1]$ is well-ordered.
\item
$x=\frac{p}{1+p}$ for $p\in \AAA\cap[\frac{1}{n+1}, \frac{1}{n}]$
and $p$ is not a limit element.

We take the segment $[\frac{p}{1+p}, \frac{r}{1+r}]$
obtained in the first step of the splitting and
the corresponding sequence $r_0, r_1, \ldots$.
$\frac{p}{1+p}$ is the limit of $r_0, r_1, \ldots$.
$[\frac{p}{1+p}, 1]$ is well-ordered because
each $[r_i, 1]$ is well-ordered.
\item
$x\neq\frac{p}{1+p}$ for any $p\in \AAA\cap[\frac{1}{n+1}, \frac{1}{n}]$.
Then, $x\neq r_0$ because $r_0=\frac{r}{1+r}$ 
for $r\in \AAA\cap[\frac{1}{n+1}, \frac{1}{n}]$.
Hence, $x=r_{i+1}$ for some $i\geq 0$.

$\AAA \cap [r_i, 1]$ is well-ordered because $r_{i+1}<r_i$.
Hence, it is enough to prove that 
$\AAA\cap [r_{i+1}, r_i]$ is well-ordered.

For a contradiction, assume that $\AAA\cap [r_{i+1}, r_i]$
contains an infinite monotonically increasing sequence $x_1, x_2, \ldots$.

\begin{Claim}
\label{MonotonousSubsequence}
Let
$x_1\in \AAA\cap [r_{i+1}, r_i]$, $x_2\in \AAA\cap [r_{i+1}, r_i]$, $\ldots$.
There is an $s\in\bbbn$ and sequences $x'_1, x'_2, \ldots$ and
$p_{j, 1}, p_{j, 2}, \ldots$ for $j\in\{1, \ldots, s\}$ such that
\begin{enumerate}
\item
$x'_1, x'_2, \ldots$ is a subsequence of $x_1, x_2, \ldots$,
\item
$x'_k\in \AAA$ follows from $p_{1, k}, \ldots, p_{s, k}\in\AAA$
and the rule \ref{Rule}, and
\item
$p_{j, 1}= p_{j, 2}= \ldots$ or $p_{j, 1}>p_{j,2}>\ldots$
for all $j\in\{1, \ldots, s\}$.
\end{enumerate}
\end{Claim}

\noindent
{\bf Proof.}
Denote
\[ s_1=\left\lceil \frac{x}{\frac{x}{r_{i+1}}+x-1} \right\rceil .\]
Consider the applications of the rule \ref{Rule} that prove 
$r_1\in \AAA, r_2\in \AAA, \ldots$.
By Lemma \ref{p1Quantity},
\[ s \leq \frac{x}{\frac{x}{p}+x-1} \leq \frac{x}{\frac{x}{r_{i+1}}+x-1}\leq s_1 \]
in each of these applications.
Hence, there exists an $s_0\in \{1, \ldots, s_1\}$ such that
infinitely many of $x_1, \ldots$ are generated by applications
of the rule \ref{Rule} with $s=s_0$.
We denote this subsequence $x_1^{(0)}, x_2^{(0)}, \ldots$.

Next, we select $x_1^{(1)}, x_2^{(1)}, \ldots$,
a subsequence of $x_1^{(0)}, x_2^{(0)}, \ldots$.
Then, we select $x_1^{(2)}, x_2^{(2)}, \ldots$,
a subsequence of $x_1^{(1)}, x_2^{(1)}, \ldots$.
We continue so until we obtain $x_1^{(s_0)}, x_2^{(s_0)}, \ldots$.

The subsequence $x_1^{(k)}, x_2^{(k)}, \ldots$ is generated from
$x_1^{(k-1)}, x_2^{(k-1)}, \ldots$ as follows:

Let $p_{1, j}^{(k-1)}, \ldots, p_{s_0, j}^{(k-1)}$
be the values of $p_1, \ldots, p_{s_0}$ in the application of
the rule \ref{Rule} that proves $x_j^{(k-1)}\in \AAA$.
We use the infinite version of Dilworth's lemma.

\begin{Theorem}
Let $y_1, y_2, \ldots$ be a sequence of real numbers.
Then $y_1$, $y_2$, $\ldots$ contains
\begin{itemize}
\item
a subsequence $y_{n_1}, y_{n_2}, \ldots$ 
such that $y_{n_1}=y_{n_2}=\ldots$, or
\item
an infinite monotonically increasing subsequence, or
\item
an infinite monotonically decreasing subsequence.
\end{itemize}
\end{Theorem}

The sequence $p_{k, 1}^{(k-1)}, p_{k, 2}^{(k-1)}, \ldots$
does not contain an infinite monotonically increasing subsequence
because all elements of this sequence belong to $\AAA\cap[r_i, 1]$
and $\AAA\cap[r_i, 1]$ is well-ordered in decreasing ordering.
Hence, this sequence contains an infinite subsequence consisting of
equal elements or an infinite monotonically decreasing subsequence.

Let this subsequence be $p_{k, n_1}^{(k-1)}$, $p_{k, n_2}^{(k-1)}$, $\ldots$.
We choose $r_{n_1}^{(k-1)}, r_{n_2}^{(k-1)}, \ldots$ as 
the sequence $x_1^{(k)}, x_2^{(k)},\ldots$.

$x_1^{(s_0)}, x_2^{(s_0)}, \ldots$
is the needed sequence $x'_1, x'_2, \ldots$.
We have
\[ p_{1, k}=p_{2, k}=\ldots \mbox{ or } 
p_{1, k}>p_{2, k}>\ldots \]
because such property holds for the sequence
$x_1^{(k)}, x_2^{(k)}, \ldots$ and $x_1^{(s_0)}$, $x_2^{(s_0)}$, $\ldots$
is a subsequence of $x_1^{(k)}$, $x_2^{(k)}$, $\ldots$.
$\Box$

We have 
\[ p_{1, 1}\geq p_{2, 1}\geq \ldots \]
\[ \ldots \]
\[ p_{1, s}\geq p_{2, s}\geq \ldots .\]
By Lemma \ref{Monotonicity}, 
\[ x'_1\geq x'_2\geq x'_3\ldots .\]
Hence, $x_1, x_2, \ldots$ 
contains an infinite non-increasing subsequence. 
$\Box$

This is a contradiction with the assumption that $x_1$, $x_2$, $\ldots$ is
monotonically increasing.
\end{enumerate}
$\Box$

Next, we construct a system of notations $S$ 
for $\AAA\cap [\frac{1}{n+1}, \frac{1}{n}]$.
We start with technical results necessary for our construction.
In section \ref{subsub:distinguishing}, we show how to distinguish
limit elements from successor elements.
In section \ref{subsub:xdminimal}, we define 
$(x, d)$-minimal sets and show that such sets can
be computed algorithmically.
Finally, in section \ref{subsub:system},
we use these results to construct a system of notations.

\subsubsection{Distinguishing elements of different types}
\label{subsub:distinguishing}

The maximal element of the set $\AAA$ is $1$.
It does not belong to $\AAA\cap [r_{i+1}, r_i]$.
Hence, $\AAA\cap [r_{i+1}, r_i]$ does not contain the maximal element
and, constructing a system of notations, we should distinguish 
numbers $p$ of three types:
\begin{enumerate}
\item
$p\in \AAA\cap [r_{i+1}, r_i]$ and $p$ is a successor. 
Then $k_S(p)=1$.
\item
$p\in \AAA\cap [r_{i+1}, r_i]$ and $p$ is a limit element. 
Then $k_S(p)=2$.      
\item
$p\notin \AAA\cap [r_{i+1}, r_i]$. 
Then $k_S(p)=3$.
\end{enumerate}
Two lemmas below shows how to distinguish 
between limit and successor elements.

\begin{Lemma}
\label{Limit}
Let $x\in \AAA\cap [r_{i+1}, r_i]$.
Then $x$ is a limit element if and only if it can
be generated by rule \ref{Rule} so that at least one of 
$p_1, \ldots, p_s$ is limit element.
\end{Lemma}

\noindent
{\bf Proof.}

``if'' part.
Assume that $p_j$ is a limit element.
Let $p_{j, 1}, p_{j, 2}, \ldots$ be a monotonically decreasing
sequence converging to $p_j$ and 
$x_k$ be the number generated by the application
of the rule \ref{Rule} to $p_1$, $\ldots$,
$p_{j-1}$, $p_{j, k}$, $p_{j+1}$, $\ldots$, $p_s$.
Then, $x_1, x_2, \ldots$ is a monotonically decreasing sequence 
converging to $x$.
Hence, $x$ is a limit element.

``only if'' part.
Let $x$ be a limit element and 
$x_1, x_2, \ldots$ be a monotonically decreasing sequence
converging to $x$.
We apply Claim \ref{MonotonousSubsequence} to $x_1, x_2, \ldots$
and obtain a subsequence $x'_1, x'_2, \ldots$.

We consider the sequences $p_{j, 1}, p_{j, 2}, \ldots$.
Let
\[ p'_j=\lim_{k\rightarrow\infty}p_{j, k} .\]
By Lemma \ref{Limits},
$x$ can be generated from $p'_1, p'_2, \ldots, p'_{s}$ 
by an application of rule \ref{Rule}.
We have
\[ p_{j, 1}=p_{j, 2}=\ldots \mbox{ or } p_{j, 1}>p_{j, 2}>\ldots .\]
for any $j\in\{1, \ldots, m\}$.
If $p_{j, 1}=p_{j, 2}=\ldots$ for all $j$,
then, $x'_1=x'_2=\ldots$.
A contradiction with the assumption
that $x_1, x_2, \ldots$ is monotonically decreasing.

Hence, 
\[ p_{j, 1}>p_{j, 2}>\ldots \]
for at least one $j$ and
$p'_j=\lim_{k\rightarrow\infty}p_{j, k}$ is a limit element.
$\Box$

\begin{Lemma}
\label{An}
Let $x\in A_n$.
Then $x$ is a limit element.
\end{Lemma}

\noindent
{\bf Proof.}
We have three cases.
\begin{enumerate}
\item
$x=\frac{p}{1+p}$ for $p\in \AAA\cap[\frac{1}{n+1},\frac{1}{n}]$
and $p$ is a limit element.

Let $p$ be the limit of $p_1, p_2, \ldots$.
Then, $\frac{p}{1+p}$ is the limit of
$\frac{p_1}{1+p_1}, \frac{p_2}{1+p_2}, \ldots$
because the function $\frac{x}{1+x}$ is continuous.
\item
$x=\frac{p}{1+p}$ for $p\in \AAA\cap[\frac{1}{n+1}, \frac{1}{n}]$
and $p$ is not a limit element.

We take the segment $[\frac{p}{1+p}, \frac{r}{1+r}]$
obtained in the first step of the splitting and
the corresponding sequence $r_0, r_1, \ldots$.
$\frac{p}{1+p}$ is the limit of $r_0, r_1, \ldots$.
\item
$x\neq\frac{p}{1+p}$ for any $p\in \AAA\cap[\frac{1}{n+1}, \frac{1}{n}]$.

Then, $x=r_i$.
We prove the lemma by induction over $i$.

\noindent
{\em Base Case.}
If $i=0$, then $r_i=\frac{r}{1+r}$ and we already
know that $\frac{r}{1+r}$ is a limit element.

\noindent
{\em Inductive Case.}
Lemma \ref{Formula} and the definition of $r_{i+1}$ imply
that $r_{i+1}\in \AAA$ follows from $r_i\in \AAA$
and $p\in \AAA$ by the rule \ref{Rule}.
If $r_i$ is a limit element, then, by Lemma \ref{Limit},
$r_{i+1}$ is a limit element, too.
\end{enumerate}
$\Box$

\subsubsection{$(x, d)$-minimal sets}
\label{subsub:xdminimal}

In the algorithms of subsection \ref{subsub:system}, 
we will often need to compute the largest element of $\AAA\cap [r_{i+1}, r_i]$
which is less than some given $x$.
This will be done by checking 
$p_1\in \AAA\cap[r_i, 1]$, $p_2\in \AAA\cap[r_i, 1]$, $\ldots$, 
$p_s\in \AAA\cap[r_i, 1]$ that can generate $x\in \AAA$ by rule \ref{Rule}.
There are infinitely many possible combinations of $p_1$, $\ldots$, $p_s$.
Hence, we need
\begin{itemize}
\item
to prove that it is enough to check finitely many combinations
$p_1\in \AAA\cap[r_i, 1]$, $p_2\in \AAA\cap[r_i, 1]$, $\ldots$, 
$p_s\in \AAA\cap[r_i, 1]$, and
\item
to construct an algorithm finding the list of combinations
$p_1\in \AAA\cap[r_i, 1]$, $p_2\in \AAA\cap[r_i, 1]$, 
$\ldots$, $p_s\in \AAA\cap[r_i, 1]$ which must be checked 
when the functions $k_S, p_S, q_S$ are computed.
\end{itemize}
We do it below.
First, we give formal definitions.

\begin{Definition}
A tuple $\langle p_1, \ldots, p_s\rangle$ is said to be $(x,d)$-allowed
if $p_1\in \AAA\cap[r_i, 1]$, $\ldots$, $p_s\in \AAA\cap[r_i, 1]$ and
$\sum_{j=1}^{s}(\frac{x}{p_j}+x-1)\leq d$.
\end{Definition}

\begin{Definition}
A tuple $\langle p_1, \ldots, p_s\rangle$ is said to be less than
or equal to $\langle p'_1, \ldots, p'_s\rangle$ 
if $p_1\leq p'_1$, $\ldots$, $p_s\leq p'_s$.
\end{Definition}

\begin{Definition}
A set of tuples $P$ is said to be $(x, d)$-minimal if,
\begin{enumerate}
\item
It contains only $(x, d)$-allowed tuples;
\item
For each $(x, d)$-allowed tuple 
$\langle p_1, \ldots, p_s \rangle$
there is a tuple belonging to $P$
which is less than or equal to $\langle p_1, \ldots, p_s \rangle$
\end{enumerate}
\end{Definition}

Next three lemmas show why $(x, d)$-allowed tuples 
and $(x, d)$-minimal sets are important for our construction.

\begin{Lemma}
\label{AllowedTuples}
$\langle p_1, \ldots, p_s\rangle$ is $(x, x)$-allowed if and only if
the application of the rule \ref{Rule} to $p_1, \ldots, p_s$
generates a number $p$ satisfying $p\geq x$.
\end{Lemma}

\noindent
{\bf Proof.}
Let $d=\sum_{j=1}^s (x+\frac{x}{p_j}-1)$.
$\langle p_1, \ldots, p_s\rangle$ is $(x, x)$-allowed if and only if $d\leq x$.
Hence, it is enough to prove that $d\leq x$ if and only if $x\leq p$.
\[ d=\sum_{j=1}^s \left(x+\frac{x}{p_j}-1\right)
=\sum_{j=1}^s \left(x+\frac{x}{p_j}-1\right)-p+p\]
\[ =\sum_{j=1}^s\left(x+\frac{x}{p_j}-1\right)
-\sum_{j=1}^s\left(p+\frac{p}{p_j}-1\right)+p=
 \left(\sum_{j=1}^s\left(1+\frac{1}{p_j}\right)\right)(x-p)+p .\]
We have 
\[ \sum_{j=1}^s\left(1+\frac{1}{p_j}\right)\geq 1+\frac{1}{p_j} >1 .\]
Hence, if $x>p$,then $(x-p)>0$ and $d>(x-p)+p=x$.
If $x\leq p$, then $(x-p)\leq 0$ and $d\leq (x-p)+p=x$.
$\Box$

\begin{Lemma}
\label{MinimalSetLemma}
Let $P$ be a $(x, x)$-minimal set.
Then, for any $p_1, \ldots, p_s$
that generates $p\geq x$ by an application of the rule \ref{Rule},
there exists a tuple $\langle p'_1, \ldots, p'_s\rangle\in S$
such that $p'_1\leq p_1$, $\ldots$, $p'_s\leq p_s$.
\end{Lemma}

\noindent
{\bf Proof.}
By Lemma \ref{AllowedTuples}, 
$\langle p_1, \ldots, p_s\rangle$ is $(x,x)$-allowed. 
By the definition of $(x,x)$-minimal set, $P$ contains
a tuple $\langle p'_1, \ldots, p'_s\rangle$ such that
$p'_1\leq p_1$, $\ldots$, $p'_s\leq p_s$.
$\Box$

\begin{Lemma}
\label{ExactCombination}
Let $P$ be a $(x, x)$-minimal set, 
$p_1\in \AAA\cap[r_i, 1], \ldots, p_s\in \AAA\cap[r_i, 1]$. 
If $x\in \AAA$ follows from $p_1, \ldots, p_s\in \AAA$ and the rule \ref{Rule},
then $\langle p_1, \ldots, p_s\rangle\in P$.
\end{Lemma}

\noindent
{\bf Proof.}
By Lemma \ref{AllowedTuples}, $\langle p_1, \ldots, p_s\rangle$
is $(x, x)$-allowed.
Hence, by Lemma \ref{MinimalSetLemma},
there exists $(x, x)$-allowed
$\langle p'_1, \ldots, p'_s\rangle\in P$
such that $p'_1\leq p_1, \ldots, p'_s\leq p_s$.

Let $x'$ be the number generated by an application of
the rule \ref{Rule} to $p'_1\in \AAA$, $\ldots$, $p'_s\in \AAA$.
If $p'_j<p_j$ for some $i$, then $x'<x$ (Lemma \ref{Monotonicity})
and $\langle p'_1, \ldots, p'_s\rangle$ is not $(x, x)$-allowed
(Lemma \ref{AllowedTuples}).

However, $(x, x)$-allowed set contains only 
$(x, x)$-allowed tuples.
Hence, $p_1=p'_1, \ldots, p_s=p'_s$, i.e.
$\langle p_1, \ldots, p_s\rangle\in P$.
$\Box$

Next lemma shows that $(x, d)$-minimal sets can be
computed algorithmically.
Its proof also shows that
a finite $(x, d)$-minimal set always exists.

\begin{Lemma}
\label{lem:xdminimal}
Assume that a system of notations for $\AAA\cap[r_i, 1]$ is given.
There is an algorithm $xdminimal(x, d)$ which receives
$x\in \AAA\cap [r_{i+1}, r_i]$ and $d\in[0, x]$ and
returns a $(x, d)$-minimal set.
\end{Lemma}

\noindent
{\bf Proof.}
We use an auxiliary procedure $findsmallest(P, x, d)$.
It receives numbers $x, d$ and an $(x, d)$-minimal set $P$
and returns the smallest $d'$ such that 
$d'>d$ and $\sum_{i=1}^s (\frac{x}{p_i}+x-1)=d'$
for some $p_1, \ldots, p_s\in \AAA$.

Both $findsmallest$ and $xdminimal$ use a constant $p_0$.
$p_0$ is defined as the largest number in $\AAA\cap[r_i, 1]$
such that $x+\frac{x}{p_0}-1>0$.
Equivalently, $p_0$ is the number in $\AAA\cap[r_i, 1]$
with the smallest $x+\frac{x}{p_0}-1$
such that $x+\frac{x}{p_0}-1>0$.
$\Delta$ denotes $\frac{x}{p_0}+x-1$.

{\em Algorithm }$findsmallest(P, x, d)$:
\begin{enumerate}
\item
Let $d'=1$;
\item
For each $\langle p_1, \ldots, p_s\rangle\in P$ do:
\begin{enumerate}
\item
For each $j\in\{1, \ldots, s\}$ :
\begin{enumerate}
\item
Find $p'_j=\max\{ p |p\in \AAA\cap[r_i, 1] \mbox{ and } p<p_j\}$,
using the given system of notations for $\AAA\cap [r_i, 1]$.
\item
$d_1=\sum_{k=1}^{j-1} (\frac{x}{p_k}+x-1)+(\frac{x}{p'_j}+x-1)+
  \sum_{k=j+1}^{s} (\frac{x}{p_k}+x-1)$.
If $d_1>d$, then $d'=\min(d', d_1)$.
\end{enumerate}
\item
$d_2=\sum_{j=1}^{s} (\frac{x}{p_j}+x-1)+(\frac{x}{p_0}+x-1)$;
If $d_2>d$, then $d'=\min(d', d_2)$.
\end{enumerate}
\item
Return $d'$ as the result;
\end{enumerate}

{\em Algorithm }$xdminimal(x, d)$
\begin{enumerate}
\item
Let $P=\emptyset$;
\item
If $d<\Delta$, return the empty set as the result;
\item
Let $y$ be the smallest number in $\AAA\cap [r_i, 1]$
such that $\frac{x}{y}+x-1<d$.
\item
while ($\frac{x}{y}+x-1 >0$) do:
\begin{enumerate}
\item
$d'=d-(\frac{x}{y}+x-1)$;
\item
$P_1=xdminimal(x, d')$;
\item
If $P_1=\emptyset$, add $\langle y\rangle$ to $P$.
Otherwise, for each $\langle p_1, \ldots, p_s\rangle\in P_1$, 
add $\langle y, p_1, \ldots, p_s\rangle$ to $P$;
\item
Replace $y$ by a greater element of $\AAA\cap[r_i, 1]$:
\begin{enumerate}
\item
If $y$ is a successor element, replace $y$ by $p_{S_1}(y)$,
using the given system of notations for $\AAA\cap [r_i, 1]$;
\item
If $y$ is a limit element, replace $y$ by $y'$ where $y'$
is the smallest element of $\AAA\cap[r_i, 1]$ such that 
\[ \frac{x}{y'}+x-1\leq d-findsmallest(P_1, x, d'). \]
\end{enumerate}
\end{enumerate}
\item
Return $P$.
\end{enumerate}

\noindent
{\em Proof of correctness for $xdminimal(x, d)$.}
We prove the correctness by 
induction over $\lfloor \frac{d}{\Delta} \rfloor$.

\noindent
{\em Base Case.}
$d\in [0, \Delta[$.

Then, $\frac{x}{y}+x-1\geq\Delta$ for any $y$.
Hence, $\sum_{j=1}^{s}(\frac{x}{p_j}+x-1)\geq\Delta$
for any $\langle p_1, \ldots, p_s\rangle$ and
there are no $(x, d)$-allowed tuples.
In this case, the algorithm returns the empty set.
Hence, it works correctly.

\noindent
{\em Inductive Case.}
We assume that the lemma holds for $d\in [0, k\Delta[$
and prove it for $d\in [k\Delta, (k+1)\Delta[$.
We use

\begin{Claim}
\label{RecursiveCall}
If $xdminimal(x, d)$ calls $xdminimal(x, d')$, then
$d'\leq d-\Delta$
\end{Claim}

\noindent
{\bf Proof.}
From the description of $xdminimal$ we have $d'=d-(\frac{x}{y}+x-1)$.
By definition of $p_0$ and $\Delta$,
$\frac{x}{y}+x-1\geq\Delta$ and $d'\leq d-\Delta$.
$\Box$

Hence, $xdminimal(x, d)$ 
calls only $xdminimal(x, d')$
with $d'<(k+1)\Delta-\Delta=k\Delta$.
The correctness $xdminimal(x, d')$ for such values
follows from the inductive assumption.

First, we prove that the computation of $xdminimal(x, d)$
always terminates.
Each $xdminimal(x, d')$ called by $xdminimal(x, d)$ terminates
because $xdminimal(x, d')$ is correct.
Hence, each while loop terminates and,
if $xdminimal(x, d)$ does not stop then
while loop is executed infinitely many times.

Let $y_j$ be the value of $y$ during
the $j^{\rm th}$-th execution of while loop.
$y$ is increased at the end of each while loop.
Hence, $y_1<y_2<\ldots$.

$y_1\in \AAA\cap [r_i, 1], y_2\in \AAA\cap [r_i, 1], \ldots$.
If while loop is executed infinitely many times, then
$y_1, y_2 \ldots$ is an infinite monotonically increasing sequence.
However, $\AAA\cap[r_i, 1]$ does not contain such sequences 
because it is well-ordered.

Hence, while loop is executed finitely many times and
$xdminimal(x, d)$ terminates.
Let $P=xdminimal(x, d)$.
Next, we prove that $P$ is a $(x, d)$-minimal set.

For a contradiction, assume that it is not.
Then, there exists an $(x, d)$-allowed tuple 
$\langle p_1, \ldots, p_s\rangle$ such that
$P$ does not contain any tuple that is less than or equal to
$\langle p_1, \ldots, p_s\rangle$.

We assume that $\langle p'_1, p_2, \ldots, p_s\rangle$
is not $(x, d)$-allowed for any $p'_1\in \AAA\cap[r_i, 1]$
satisfying $p'_1<p_1$.
(Otherwise, we can replace $p_1$ by the smallest 
$p'_1\in \AAA\cap[r_i, 1]$ such that $\langle p'_1, p_2, \ldots, p_s\rangle$
is $(x, d)$-allowed.)

Consider two cases:
\begin{enumerate}
\item
In $xdminimal(x, d)$, while loop is executed with $y=p_1$.

Denote $d'=d-(\frac{x}{p_1}+x-1)$.
The tuple $\langle p_2, \ldots, p_s\rangle$ is $(x, d')$-allowed.

$xdminimal(x, d)$ calls $xdminimal(x, d')$.
$xdminimal(x, d')$ works correctly, i.e. returns
an $(x, d')$-minimal set $P_1$.
Hence, $P_1$ contains a tuple $\langle p'_2, \ldots, p'_s\rangle$
that is less than or equal to $\langle p_2, \ldots, p_s\rangle$.

$xdminimal(x, d)$ adds 
$\langle p_1, p'_2, \ldots, p'_s\rangle$ to $P$
because $\langle p'_2, \ldots, p'_s\rangle$
belongs to the set returned by $xdminimal(x, d')$.
Hence, $P$ contains the tuple $\langle p_1, p'_2, \ldots, p'_s\rangle$
that is less than or equal to $\langle p_1, p_2, \ldots, p_s\rangle$.
A contradiction.
\item
While loop is not executed with $y=p_1$.

Let $y_1$ be the greatest number such that 
$y_1<p_1$ and while loop is executed with $y=y_1$.
Let $y_2$ be the number by which $y_2$ is replaced
in the end of while loop.

$y_1<y_2$ because $y$ is always replaced by a greater number. 
By definition of $y_1$, $y_2>p_1$.
(Otherwise $y_2$ would have been instead of $y_1$.)
\begin{enumerate}
\item
$y_1$ is a successor element.

Then, $y_1$, $y_2$, $p_1$ all belong to $\AAA$ and $y_1<p_1<y_2$.
When $xdminimal(x, d)$ replaces 
$y_1$ by a greater element of $\AAA$, it chooses
the smallest element of $\AAA$ that is greater than $y_1$.
It can be $p_1$ or some number between $y_1$ and $p_1$
but not $y_2$. 
A contradiction.
\item
$y_1$ is a limit element.

We assumed that $\langle p'_1, p_2, \ldots, p_s\rangle$
is not $(x, d)$-allowed for any
$p'_1\in \AAA\cap[r_i, 1]$ satisfying $p'_1<p_1$.
Hence, $\langle y_1, p_2, \ldots, p_s\rangle$ is 
not $(x, d)$-allowed i.e.
\[ \left(\frac{x}{y_1}+x-1\right)+
\sum_{j=2}^s\left(\frac{x}{p_j}+x-1\right) >d .\]
\[ \sum_{j=2}^s\left(\frac{x}{p_j}+x-1\right)
>d-\left(\frac{x}{y_1}+x-1\right)=d' \]
Hence, 
\[ \sum_{j=2}^s\left(\frac{x}{p_j}+x-1\right)
\geq findsmallest(P_1, x, d') \]
where $P_1$ is the $(x, d')$-minimal set obtained
by $xdminimal(x, d')$.
However,
\[ \sum_{j=1}^s\left(\frac{x}{p_j}+x-1\right)\leq d \]
because $\langle p_1, p_2, \ldots, p_s\rangle$ is 
$(x, d)$-allowed.
Hence,
\[ \frac{x}{p_1}+x-1\leq d-\sum_{j=2}^s\left(\frac{x}{p_j}+x-1\right) \leq 
  d-findsmallest(P_1, x, d') .\]
By the definition, $y_2$ is the smallest number such that
\[ \frac{x}{y_2}+x-1\leq d-findsmallest(P_1, x, d') .\] 
This implies $y_2\leq p_1$.
A contradiction with $y_2>p_1$.
\end{enumerate}
\end{enumerate}
$\Box$

\subsubsection{System of notations}
\label{subsub:system}

We show how to extend a system of notations
$S$ from $\AAA\cap[\frac{1}{n},1]$ to $\AAA\cap[\frac{1}{n+1}, 1]$.
Below, we give the algorithms computing $k_S(x)$, $p_S(x)$ and $q_S(x)$
for $x\in[\frac{1}{n+1},\frac{1}{n}]$.
These algorithms use the procedure $xdminimal(x, d)$ defined in 
the previous subsection.
They also use the system $S$ for $\AAA\cap[\frac{1}{n},1]$.

{\em Function $k_S(x)$.}
\begin{enumerate}
\item
Use the system for $\AAA\cap[\frac{1}{n},\frac{1}{n-1}]$
to find whether $x=\frac{p}{1+p}$ for some
$p\in \AAA\cap[\frac{1}{n},\frac{1}{n-1}]$.
If yes, then $k_S(x)=2$.
\item
Otherwise, find the segments $[\frac{p}{1+p}, \frac{r}{1+r}]$
and $[r_{i+1}, r_i]$ containing $x$.
If $x=r_{i+1}$ or $x=r_i$, then $k_S(x)=2$.
\item
Otherwise,
find an $(x, x)$-minimal set $P$ using $xdminimal(x, x)$.
\item
If there exists $\langle p_1, \ldots, p_s\rangle\in P$ such that
$x$ is generated by an application of the rule \ref{Rule} to
$p_1, \ldots, p_s$ and at least one of $p_1, \ldots, p_s$
is a limit element, then $k_S(x)=2$.
\item
Otherwise, if there exists $\langle p_1, \ldots, p_s\rangle\in P$ such that
$x$ is generated by an application of the rule \ref{Rule} to
$p_1, \ldots, p_s$, then $k_S(x)=1$.
\item
Otherwise, $k_S(x)=3$.
\end{enumerate}

{\em Function $p_S(x)$.}
\begin{enumerate}
\item
Find the interval $[r_{i+1}, r_i]$ containing $x$.
Execute $xdminimal(x, x)$ and find
a $(x, x)$-minimal set.
\item
Let $P_1$ be the set consisting of
all tuples $\langle p_1, \ldots, p_s\rangle$ such that 
\begin{enumerate}
\item
$\langle p_1, \ldots, p_s\rangle\in P$ or
\item
$\langle p_1, \ldots, p_{j-1}, p'_j, p_{j+1}, \ldots, p_s\rangle\in P$ 
and $p_j=p_S(p'_j)$ for some $j\in\{1, \ldots, s\}$ or
\item
$\langle p_1, \ldots, p_{j-1}, p', p_{j+1}, \ldots, p_s\rangle\in P$ for
some $j\in\{1, \ldots, s\}$ and $p'\in \AAA\cap[r_i, 1]$.
\end{enumerate}
\item
For each tuple $\langle p_1, \ldots, p_s\rangle\in P_1$
find the number $p\in \AAA$ generated by an application
of the rule \ref{Rule} to $p_1, \ldots, p_s$.

$p_S(x)$ is the smallest of those $p$ which are greater than $x$.
\end{enumerate}

{\em Function $q_S(x)$.}
\begin{enumerate}
\item
If $x=\frac{p}{1+p}$, $p\in \AAA\cap[\frac{1}{n},\frac{1}{n-1}]$
and $p$ is a limit element,
$q_S(x)$ is a program computing
$\frac{\varphi_{q_S(p)}(0)}{1+\varphi_{q_S(p)}(0)}$,
$\frac{\varphi_{q_S(p)}(1)}{1+\varphi_{q_S(p)}(1)}$,
$\ldots$.
\item
If $x=\frac{p}{1+p}$, $p\in \AAA\cap[\frac{1}{n},\frac{1}{n-1}]$
and $p$ is a successor element, find $r=p_S(p)$.
$q_S(x)$ is a program computing the sequence
$r_0$, $r_1$, $\ldots$ corresponding to
$[\frac{p}{1+p}, \frac{r}{1+r}]$.
\item
Otherwise, search the set $P$ returned by $xdminimal(x, x)$ and
find $p_1\in \AAA\cap [r_i,1]$,
$\ldots$, $p_s\in \AAA\cap [r_i,1]$ such that 
$x$ is generated by an application of the rule \ref{Rule} to
$p_1, \ldots, p_s$ and $p_j$ is a limit element.

$q_S(x)$ is a program computing 
the sequence $x_1$, $x_2$, $\ldots$ where $x_k$ is 
generated by an application of the rule \ref{Rule} to
$p_1$, $\ldots$, $p_{j-1}$,
$\varphi_{q_{S_1}(p_j)}(k)$, $p_{j+1}$, $\ldots$, $p_s$.
\end{enumerate}

\begin{Lemma}
\label{lem:system}
$S$ is a system of notations for $\AAA\cap[\frac{1}{n+1},1]$.
\end{Lemma}

\noindent
{\bf Proof.}
By transfinite induction over $A_n$.

\noindent
{\em Base Case.}
$S$ is a correct system of notations
for $\AAA\cap[\frac{1}{n},1]$.

\noindent
{\em Inductive Case.}
Let $y\in A_n$.
We assume that $S$ is correct for all $\AAA\cap[y', 1]$
with $y'\in A_n$ and 
$y'>y$ and prove that it is correct for $\AAA\cap[y, 1]$.
We consider two cases:
\begin{enumerate}
\item
$y=\frac{p}{1+p}$ and $p\in \AAA\cap[\frac{1}{n},\frac{1}{n-1}]$.

Similarly to the proof of Lemma \ref{An},
$y$ is a limit of a sequence consisting of elements of $A_n$.
Hence, if $x>y$, then $x>y'$ where $y'$ is some element of
this sequence.
The functions $k_S(x)$, $p_S(x)$, $q_S(x)$
are correct because $S$ is correct for $\AAA\cap[y', 1]$
(by inductive assumption).
It remains to prove the correctness of
$k_S(x)$, $p_S(x)$, $q_S(x)$ for $x=y$.

$k_S(y)=2$.
This is correct because,
by Lemma \ref{An}, $y$ is a limit element.
The function $p_S(x)$ is defined only for successor elements.
Hence, we do not need to check its correctness for the limit element $y$.
The correctness of the sequence computed by
$q_S(y)$ is proved in the proof of Lemma \ref{An}.

\item
$y=r_{i+1}$ for $i\geq 0$.
In this case, we assume that $S$ is correct for $\AAA\cap[r_i, 1]$
and prove the correctness for $\AAA\cap[r_{i+1}, r_i]$.

By Lemma \ref{lem:xdminimal}, $xdminimal(x, d)$
returns an $(x, d)$-minimal set if it has access
to a system of notations for $\AAA\cap[r_i, 1]$.
We know that $S$ is correct for $\AAA\cap[r_i, 1]$.
Hence, the set $P$ returned by $xdminimal(x, x)$
is $(x, x)$-minimal.

\begin{enumerate}
\item[2.1.]
Proof of correctness for $k_S$.

If $x\in \AAA\cap[r_{i+1}, r_i]$, then
$x\in \AAA$ follows from $p_1\in \AAA, \ldots, p_s\in \AAA$ and the rule \ref{Rule},
for some $p_1, \ldots, p_s$.
By Lemma \ref{p1Second},
$p_1\in \AAA\cap[r_i,1], \ldots, p_s\in \AAA\cap[r_i,1]$.
By Lemma \ref{ExactCombination},
$\langle p_1, \ldots, p_s\rangle$ belongs to $P$.

Correctness of $xdmininal(x, x)$ implies that, 
if $x\in \AAA$, the algorithm computing $k_S$ finds
$p_1, \ldots, p_s$ such that $p\in \AAA$ follows from
$p_1\in \AAA, \ldots, p_s\in \AAA$ and the rule \ref{Rule}.

Hence, it distinguishes $x\in \AAA$ and $x\notin \AAA$ correctly.
By Lemma \ref{Limit}, it distinguishes
limit and successor elements correctly.

\item[2.2.]
Proof of correctness for $p_S$.

We prove that $p_S(x)$ returns the element of $\AAA\cap[r_{i+1}, r_i]$
immediately preceding $x$ i.e. 
$(\forall z\in \AAA\cap[r_{i+1}, r_i])(x<z \Rightarrow p_S(x)\leq z)$.

Let $z\in \AAA\cap[r_{i+1}, r_i]$ and $x<z$.
Consider $p_1, \ldots, p_s$ that generate $z\in \AAA$ by rule \ref{Rule}.

$P$ contains a tuple 
$\langle p'_1, \ldots, p'_s\rangle$ such that
$p'_1\leq p_1, \ldots, p'_s\leq p_s$
(Lemma \ref{MinimalSetLemma}).
An application of the rule \ref{Rule} to 
$p'_1, \ldots, p'_s$ generates $p\in \AAA$
with $p\geq x$ (Lemma \ref{AllowedTuples}).
Consider two cases:
\begin{enumerate}
\item[(a)]
$p>x$.

The algorithm computing $p_S$ adds
$\langle p'_1, \ldots, p'_s\rangle$ to the set $P_1$.
Later, it sets $p_S(x)$ equal to a number
that is less or equal to $p$.
(This is true because $\langle p'_1, \ldots, p'_s\rangle\in P_1$
and $p'_1, \ldots, p'_s$ generates $p>x$.
The algorithm selects $p_S(x)$ as the smallest of all $p$ satisfying 
these conditions.) 

By Lemma \ref{Monotonicity}, $p\leq z$. 
Hence, $p_S(x)\leq p\leq z$.
\item[(b)]
$p=x$

If $p_1=p'_1$, $\ldots$, $p_s=p'_s$ then $p=z$.
However, $p<z$.
Hence, $p_j<p'_j$ for some $i$.
Let $p''_j=p_S(p'_j)$.
We have $p''_j\leq p_j$ because $p_S(p'_j)$ 
is the smallest element of $\AAA$ that is greater 
than $p'_j$.
Let $p$ denote the number generated by the rule \ref{Rule} from
$p'_1, \ldots, p'_{j-1}$, $p''_j$, $p'_{j+1}, \ldots, p'_s$.

By Lemma \ref{Monotonicity}, $x<p$.
Hence, the algorithm for $p_S(x)$ adds
$\langle p'_1, \ldots, p'_{j-1}$, $p''_j$, $p'_{j+1}, \ldots, p'_s\rangle$
to the set $P_1$ and, then, checking tuples in $P_1$,
sets $p_S(x)$ equal to a number which is greater than or equal to $p$.
This implies $p_S(x)\leq p$.

From $p'_1\leq p_1$, $\ldots$, $p'_{j-1}\leq p_{j-1}$,
$p''_j\leq p_j$, $p'_{j+1}\leq p_{j+1}$, $\ldots$, $p'_s\leq p_s$
it follows that $p\leq z$ (Lemma \ref{Monotonicity}).
Hence, $p_S(x)\leq p\leq z$.
\end{enumerate}
So, in both cases $p_S(x)$ is less than or equal to
any $z\in \AAA$ satisfying $x<z$.
On the other hand, $p_S(x)\in \AAA$ and $x<p_S(x)$.
(It can be seen from the algorithm computing $p_S$.)

Hence, $p_S(x)$ is the smallest element of $\AAA$ 
satisfying $x<p_S(x)$, i.e. the algorithm computes $p_S$ 
correctly.  

\item[2.3.]
Proof of correctness for $q_S$.

We already proved that, if there exist $p_1, \ldots, p_s$ 
such that $x\in \AAA$ follows from $p_1\in \AAA, \ldots, p_s\in \AAA$,
then such combination is found by $xdminimal(x, x)$
(see proof of correctness for $k_S$).
If there exists such a combination with one
of $p_1, \ldots, p_s$ being limit element,
it is found. 
The algorithm computing $q_S$ generates a program computing
required sequence from such combination correctly.
\end{enumerate}
\end{enumerate}
The correctness of $S$ for $\AAA\cap[\frac{1}{n}, 1]$
follows by transfinite induction.
$\Box$

By Lemmas \ref{InductionStep} and \ref{lem:system},
$\AAA\cap[\frac{1}{n}, 1]$ is well-ordered and has a
system of notations for any $n$.
Hence, $\AAA$ is well-ordered and has a system of notations.
This completes the proof of Theorem \ref{Basic}.
$\Box$

\subsection{Universal simulation}
\label{Subsection3}

\begin{Theorem}
\label{MainSimulation}
For any $p\in \AAA$ there exists $k$ such that
$\PFIN\langle x\rangle\subseteq [pk, k]\PFIN$ for 
all $x$ which are greater than any $p'\in \AAA\cap[0, p[$.
There exists an algorithm which receives 
a probabilistic machine $M$ and a probability $x$
and outputs a team $L_1, \ldots, L_k$ which identifies the
same set of functions.
\end{Theorem}

\noindent
{\bf Proof.}
By transfinite induction.

\noindent
{\em Base Case.}
For $p>\frac{1}{2}$, the theorem follows from the results
of \cite{Freivalds}.

\noindent
{\em Inductive Case.}
We assume that the theorem is true for all $p\in \AAA$ such that $p>p_0$
and prove it for $p=p_0$.

Let $p'_0$ be the largest element of $\AAA$
for which $\frac{x}{p'_0}+x-1>0$. 
$p'_0$ is always a successor element. 
(If it was a limit element, let $q_1, q_2, \ldots$ be a decreasing
sequence that converges to $p'_0$. For some element $q_i$ in this
sequence, $\frac{x}{q_i}+x-1>0$, implying that $p'_0$ is not the largest
element with this property.) Let $p''_0$ be the predecessor
of $p'_0$.

Let $P$ be a $(x, x)$-minimal set (see Section \ref{subsub:xdminimal}).
Let $P'$ be the set of all $p$ that appear in some tuple in the set $P$.

We define two functions $g(r)$ and $g'(r)$, for $r\in[0, x]$.
To define $g(r)$, let $y$ be the smallest element of $\AAA$ which
is at least $\frac{x}{1+x-r}$. If $y=p''$, we define $g(r)=0$.
Otherwise, let $y'$ be the largest element of $P'$ satisfying $y'\leq y$.
Let $g(r)$ be the solution to $y'=\frac{p_0}{1-p_0+g(r)}$.
(Equivalently, $g(r)=p_0+\frac{p_0}{y'}-1$.)
To define $g'(r)$, let $S(r)$ be the set of all 
tuples $\langle r_1, r_2, \ldots, r_m\rangle$ such that 
$r_1+\ldots+r_m\leq r$, $r_1>0, \ldots, r_m>0$.
Then, 
\[ g'(r)=\sup_{\langle r_1, r_2, \ldots, r_m\rangle \in S(r)}
g(r_1)+g(r_2)+\ldots+g(r_m) .\]

In the simulation algorithm for $p=p_0$, we use several simulation 
algorithms for $p>p_0$ as subroutines.
Namely, we use:

\begin{enumerate}
\item
A simulation algorithm for $p=p''_0$.
\item
Simulation algorithms for all $p\in P'$.
\end{enumerate}
The existence of these simulation algorithms is implied by the assumption
that Theorem \ref{MainSimulation} holds for $p>p_0$.

A $[pk, k]\PFIN$-team $L=\{ L_1, \ldots, L_k\}$
simulates a probabilistic $\PFIN\langle x\rangle$-machine $M$ as follows:
\begin{enumerate}
\item
$L_1, \ldots, L_k$ read $f(0)$, $f(1)$, $\ldots$,
simulate $M$ and wait until the probability 
that $M$ has issued a conjecture reaches $x$.
Then $pk$ machines ($L_1, \ldots, L_{pk}$) issue conjectures
$h_1, \ldots, h_{pk}$.
\item
The first values of the functions computed by $h_1$, $\ldots$, $h_{pk}$
are identical to the values of $f$, i.e.
\[ \varphi_{h_1}(i)=\ldots=\varphi_{h_{pk}}(i)=f(i) \]
for $i\leq m$ where $f(m)$ is the last value of $f$ read by 
$L$ before issuing conjectures.
The next values of these functions are computed as follows:

Let $n=m+1$. Let $T=\{\langle f(0), f(1), \ldots, f(m)\rangle\}$.
We repeat the following sequence of operations.
For each segment $\rho=\langle f(0), \ldots, f(n-1)\rangle$ in $T$:
\begin{enumerate}
\item
Find all conjectures of $M$ (among ones issued until the probability 
reached $x$) that output $f(0), \ldots, f(n-1)$. Run each of those 
conjectures on input $n$. Let $d_1, \ldots, d_s$ be the values
that are output by at least one of conjectures.
For $i\in\{1, \ldots, s\}$, let $r_i$ be the total
probability of $M$'s conjectures outputting $f(n)=d_i$.
The programs $h_1, \ldots, h_{pk}$ output the next value as follows.
Out of those programs, which have output the segment 
$\langle f(0), \ldots, f(n-1) \rangle$, 
$g'(r_1)k$ output $f(n)=d_1$, $g'(r_2)k$ output
$f(n)=d_2$ and so on. 
If the number of programs that have output 
the segment $\langle f(0), \ldots, f(n-1)\rangle$ is larger
than $g'(r_1)k+g'(r_2)k+\ldots+g'(r_s)k$, the remaining programs
are not necessary for the further steps. 
Make them output $f(n)=f(n+1)=\ldots=0$, 
to ensure that every program computes a total function.

\item 
\label{Splitting}
The programs which have output $f(n)=d_i$ then simulate the machine $M$
on input $f(0), \ldots, f(n)$. If the total probability of $M$ issuing 
a conjecture consistent with $f(0), \ldots, f(n)$ reaches $x$,
invoke the simulation algorithm for simulating a probabilistic
machine with the success probability $p=\frac{x}{1-x+r_i}$
by a team of $(1-p_0+g(r_i))k$ machines, $p_0 k$ of which have 
to be successful. Let $g(r_i)k$ of $g'(r_i)k$ programs which have 
output $f(0), \ldots, f(n)$ simulate the 
simulate the first $g(r_i)k$ machines in this simulation.
If $g'(r_i)k>g(r_i)k$, make the remaining $g'(r_i)k-g(r_i)k$ 
programs output $f(n+1)=f(n+2)=\ldots=0$.

\item
Otherwise (if the probability of $M$ issuing 
a conjecture consistent with $f(0), \ldots, f(n)$ does not reach $x$),
add $(f(0), \ldots, f(n))$ to the set $T$.
\end{enumerate}

After the previous three steps have 
been done for every segment $\langle f(0), \ldots, f(n-1)\rangle\in T$,
increase $n$ by 1 and repeat.

\item
After $L_1, \ldots, L_{p_0 k}$ have issued conjectures,
all remaining machines in the team $L$ read the next values of 
the input function and simulate the conjectures issued by the probabilistic
machine $M$ before conjectures of $L_1, \ldots, L_{p_0 k}$.
They wait until the step \ref{Splitting} happens, for a segment
$f(0), \ldots, f(n)$ consistent with the input.
Then,
$L_{p_0 k+1}$, $\ldots$, $L_k$ (i.e. all machines which have not
issued conjectures yet) 
participate in one of two simulations:

\begin{enumerate}
\item
\label{Large}
If $g(r_i)>0$, they, together with $g(r_i)k$ of programs 
$h_1, \ldots, h_{p_0 k}$, form an $[p_0 k, (1-p_0+g(r_i)) k]$ team.
This team simulates a probabilistic machine $M'$ according
to the algorithm for $p=\frac{p_0}{1-p_0+g(r_i)}$.
(Note that, by the definition of $g$, $p\in P'$.)  

The machine $M'$ is defined as follows. It reads 
$f(0)$, $\ldots$, $f(n)$ and then simulates $M$.
If, while reading $f(0)$, $\ldots$, $f(n)$, $M$ outputs
a conjecture inconsistent with the segment  
$f(0)$, $\ldots$, $f(n)$, $M'$ restarts the simulation of $M$.
If $M$ outputs a conjecture consistent with $f(0)$, $\ldots$, $f(n)$,
$M'$ outputs this conjecture as well. 
If $M$ outputs no conjecture while reading  
$f(0)$, $\ldots$, $f(n)$, $M'$ proceeds to read the next values
of $f$ and keeps simulating $M$.
\item
\label{Small}
If $g(r_i)=0$, they form an $[p_0 k, (1-p_0)k]$ team and simulate
the probabilistic machine $M'$, defined as above, according
to the algorithm for $p=p''_0$.
\end{enumerate}
\end{enumerate}

{\em Proof of correctness.}
We need to show two statements.
\begin{itemize}
\item
When we use an $[p_0 k, (1-p_0+g(r_i)) k]$ or an
$[p_0 k, (1-p_0)k]$ team to simulate a probabilistic machine,
the team is able to perform the simulation.
\item 
For a segment $\langle f(0), f(1), \ldots, f(n-1)\rangle\in T$,
the sum of numbers $g'(r_i)k$ of programs $h_1, \ldots, h_{pk}$
asked to output various extensions  $\langle f(0), f(1), \ldots, 
f(n-1), f(n)\rangle$ of this segment is never more than
the number of program which have output the segment
$\langle f(0), f(1), \ldots, f(n-1)\rangle$.
\end{itemize}

We start with the first statement.  
We consider two cases. 
\begin{enumerate}
\item
Case \ref{Large}.

Here, we use $g(r_i) k$ programs and $(1-p_0)k$
machines $L_{p_0 k+1}$, $\ldots$, $L_{k}$ to simulate
$M$ on functions with the given $f(0), \ldots, f(n)$ 
according to the algorithm for $p=p'_i$.

The success probability for $M'$ is at least 
$\frac{x}{1-x+r_i}$, since $M$ succeeds with probability at least $x$ 
and the probability that $M$ outputs a conjecture inconsistent 
with $f(0), \ldots, f(n)$ is $x-r_i$.
Let $y$ and $y'$ be as in the definition of $g(r_i)$.
Then, $\frac{x}{1-x+r_i}$ is greater than any $p'\in A\cap [0, y[$.
Since $y\geq y'$, $\frac{x}{1-x+r_i}$ is also 
greater than any $p'\in A\cap [0, y'[$.
Therefore, by inductive assumption, it can be simulated by an $[y'k', k']$
team, for some $k'$.
Since $y'=\frac{p_0}{1-p_0+g(r_i)}$, a $[p_0 k, (1-p_0+g(r_i))k] $ team 
can do this task, as long as $k$ is appropriately chosen. 

\item
Case \ref{Small}.

Here, we use $(1-p_0)k$
machines $L_{p_0 k+1}$, $\ldots$, $L_{k}$ to simulate
$M$ on functions with the given $f(0), \ldots, f(n)$ 
according to the algorithm for $p=p''_0$.

The success probability for $M'$ is at least 
$\frac{x}{1-x+r_i}$, by the same argument as before.
Since $g(r_i)=0$, this is more than any $p'\in A\cap [0, p''_0[$.
By inductive assumption, this means $M'$ can be simulated 
by an $[p''_0 k', k']$ team, for some $k'$.
We will choose $k$ so that $k'=(1-p_0)k$.
Then, $M'$ can be simulated by an $[p''_0(1-p_0) k, (1-p_0)k]$ 
team. It remains to prove that this simulation yields
at least $p_0 k$ correct programs. This is equivalent to
$p''_0(1-p_0) k \geq p_0 k$.

\comment{ 
 since $M$ succeeds with probability at least $x$ 
and the probability that $M$ outputs a conjecture inconsistent 
with $f(0), \ldots, f(n)$ is $x-p_i$.

The probability of conjectures consistent with input segment
becomes less than $\Delta$ and
we use the simulation algorithm for $p=p''_0$.
\begin{enumerate}
\item
The simulation is possible.

Similarly to the previous case, $M$ can be transformed into
machine $M'$ which identifies only functions consistent
with input read so far. 
The probability of success of $M'$ is equal to the success ratio of $M$.

The probability of issued conjectures is at most $\Delta$.
Hence, the success ratio is at least $\frac{x}{1-x+\Delta}=p''_0$.
Hence, $M'$ can be simulated by
a team with success ratio $p''_0$.
\item
The simulation gives $p_0 k$ correct programs.

$(1-p_0)k$ machines ($L_{p_0 k+1}, \ldots, L_k$) participate 
in this simulation. $p''_0(1-p_0)k$ of them are successful.

We have $p''_0\geq \frac{x}{1-x}$. (By the definition of
$p''_0$, it must be the case that $\frac{x}{p''_0}+x-1\leq 0$ 
which is equivalent to $p''_0\geq \frac{x}{1-x}$.)
}
If $p''_0<\frac{p_0}{1-p_0}$, 
then $\frac{p''_0}{1+p''_0}$ would belong to the interval $[x, p_0[$,
contradicting the assumption that this interval does not contain
any elements of $\AAA$.
Therefore, $p''_0(1-p_0)\geq p_0$ and $p''_0(1-p_0) k\geq p_0 k$.
\end{enumerate}

Next, we show that the programs output by $L_1$, $\ldots$, $L_{pk}$
are sufficient to conduct the necessary simulations. 
Let $\langle f(0), \ldots, f(n-1) \rangle \in T$ be an initial segment,
output by $M$ with probability $r$ and let $r_1, \ldots, r_s$ be the
probabilities of its possible extensions $\langle f(0), \ldots, 
f(n) \rangle$. Then, the number of programs $h_1$, $\ldots$, $h_{pk}$
outputting the segment $\langle f(0), \ldots, f(n-1) \rangle$ is $pk$ if 
$n=m+1$ and $g'(r)k$ if $n>m+1$.
The number of programs outputting its extensions 
$\langle f(0), \ldots, f(n) \rangle$ is
$g'(r_1)k$, $\ldots$, $g'(r_s)k$. 

For the $n>m+1$ case, it suffices to show that
$\sum_{i=1}^t g'(r_i)\leq g'(r)$, whenever $\sum_{i=1}^t r_i \leq r$. 
The $n=m+1$ case follows from the $n>m+1$ case,
once we prove $g'(x)\leq p_0$.
We now proceed to show those two results.

\begin{Lemma}
If $\sum_{i=1}^t r_i \leq r$, then $\sum_{i=1}^t g'(r_i)\leq g'(r)$.
\end{Lemma}

{\bf Proof.}
Immediate from the definition of $g'$.
$\Box$

\begin{Lemma}
$g'(x)\leq p_0$
\end{Lemma}

{\bf Proof.}
We need to prove that $g(r_1)+\ldots+g(r_m)\leq p_0$, whenever
$r_1+\ldots+r_m\leq x$ and $r_1, \ldots, r_m\geq 0$.
Let $y'_i$ be the value of $y'$ in the calculation of $g(r_i)$.

We claim that the tuple $\langle y'_1, y'_2, \ldots, y'_m\rangle$
is $(x, x)$-allowed. To prove that, we need to show 
$\sum_{i=1}^m x+\frac{x}{y'_i}-1 \leq x$. This is true
because, $y'_i \geq \frac{x}{1-x+r_i}$ and, therefore
\[ x+\frac{x}{y'_i}-1 \leq x+\frac{x}{x/(1-x+r_i)} -1 =r_i, \]
implying $\sum_{i=1}^m x+\frac{x}{y'_i}-1 \leq \sum_{i=1}^m r_i \leq x$.
Since the tuple is $(x, x)$ allowed, applying rule 2 to it
generates $p\geq x$. Since $p_0$ is the smallest element of $\AAA$
satisfying $p_0\geq x$, this also means $p\geq p_0$.

Consider the application of rule 2 to $y'_1, y'_2, \ldots, y'_m$.
Consider the values of $q_1, \ldots, q_m$ in this application.
We have $\frac{p}{q_i+1-p}=y'_i$ which is equivalent to
$q_i=p+\frac{p}{y'_i}-1$. Since $p\geq p_0$, we have
\[ q_i \geq (p-p_0) + p_0+\frac{p_0}{y'_i}-1 = (p-p_0) + g(r_i) .\]
Summing over all $i$ gives
\[ \sum_{i=1}^m g(r_i) \leq \sum_{i=1}^m q_i - m (p_0-p) = p - m (p-p_0) \]
Since $p_0\leq p$, we have $p-p_0\geq 0$ and the equation above
is at most $p-(p-p_0)=p_0$. This proves the lemma.
$\Box$

{\em The size of $L$.}
We show how to select the size of the team $L$ so that be
it will able to perform all described simulations.
Two conditions must be satisfied:
\begin{enumerate}
\item
When the machines of the team split, the amount of machines
saying that $f(m)=d_i$ must be integer for any $d_i$ i.e.,
$g'(r) k$ must be integer in all cases.
\item
When the simulation algorithm for the success ratio $p_0$
uses another simulation algorithm 
(with the ratio of successful machines $p'>p_0$),
a certain team size $k'$ is required for simulation with
$[p'k', k']\PFIN$-team.
The amount of machines participating in this simulation
(when it is used as the subroutine of the simulation
for the ratio $p_0$) must be multiple of $k'$.
\end{enumerate}

For the first condition, notice that
$g'(r)k$ is, by definition, a sum of $g(r)k$ for smaller $r$.
Therefore, it suffices to choose $k$ so that $g(r)k$ is an integer.
By definition, $g(r)=p_0+\frac{p_0}{y'}-1$ where $y'$ is belongs to
a finite set $P'$. Since $P'\subseteq \AAA$ and   
$\AAA$ is the subset of rational
numbers (section \ref{Subsection1}),
this means that $k$ must be chosen so that
$g(r)k$ is an integer for finitely many rationals $g(r)$.
Each of those requirements is equivalent to requiring
that the denominator of $g(r)$ divides $k$.

The second condition is equivalent to:
\begin{enumerate}
\item
For all $p' \in P'$, the team size
$(1-p_0+g(r))k = \frac{p_0}{p'} k$ must be a multiple of $k_i$
where $k_i$ is the size of the team with the success ratio $p'$.
\item
$(1-p_0) k$ must be a multiple of $k_0$, the
size of the simulation team with the success ratio $p'_0$.
\end{enumerate}

Overall, we have finitely many requirements.
Each of them requires that the team size is
a multiple of some finite number of integers 
$k_1, \ldots, k_m$.
If we select the size $k$ so, the simulation
algorithm will be able to perform all required simulations.
$\Box$

Theorem \ref{MainSimulation} implies

\begin{Corollary}
\label{Part2}
Let $x, y\in [0,1]$ and $x<y$.
If there is no $p\in \AAA$ satisfying $x\leq p< y$,
then
\[ \PFIN\langle x\rangle = \PFIN\langle y\rangle.\]
\end{Corollary}

{\bf Proof.}
Any machine which succeeds with probability $y$, succeeds with 
probability $x<y$, too. Hence,
it suffices to prove that any machine with
the probability of success $x$ can be simulated
by a machine with the probability of success $y$, i.e. 
\[ \PFIN\langle x\rangle\subseteq \PFIN\langle y\rangle.\]
  
Let $p$ be the smallest element of $\AAA$ which is greater 
than $x$. Theorem \ref{MainSimulation} implies
\[ \PFIN\langle x\rangle \subseteq [pk, k] \PFIN \subseteq
\PFIN\langle p\rangle.\]
We have $y\leq p$ and, hence,
\[ \PFIN\langle p\rangle\subseteq \PFIN\langle y\rangle\]  
\[ \PFIN\langle x\rangle\subseteq \PFIN\langle y\rangle\]  
$\Box$

So, if Theorem \ref{Part1} does not prove that the power of learning 
machines with probabilities $x$ and $y$ is different, then
these probabilities are equivalent. 
Hence, 

\begin{Theorem}
\label{CapabilityType}
$\AAA$ is the probability hierarchy for probabilistic $\PFIN$-type learning
in the range $[0, 1]$.
\end{Theorem}

{\bf Proof.}
Follows from Theorem \ref{Part1} and Corollary \ref{Part2}.
$\Box$

Theorem \ref{CapabilityType} has a following important corollary.

\begin{Theorem}
\label{Decidability}
Probabilistic $\PFIN$-type learning probability structure is decidable
i.e. there is an algorithm that receives as input 
two probabilities $p_1$ and $p_2$ and computes
whether $\PFIN\langle p_1\rangle=\PFIN\langle p_2\rangle$.
\end{Theorem}

{\bf Proof.}
Use the algorithm of Lemma \ref{FindInterval} to find
the intervals
$[f_1(p_1), f_2(p_1)]$ and $[f_1(p_2), f_2(p_2)]$.
If these two intervals are equal,
$\PFIN\langle p_1\rangle=\PFIN\langle p_2\rangle$.
Otherwise, $\PFIN\langle p_1\rangle\neq\PFIN\langle p_2\rangle$.
$\Box$

\section{Relative complexity}
\label{Relative}

From Theorem \ref{Basic} we know that $\PFIN$-type probability hierarchy
is well-ordered. 
A question appears: what is the ordering type
of this hierarchy?
To what particular ordinal is it order-isomorphic?
We analyze the proof of Theorem \ref{Basic} step by step.

Let $\alpha(x)$ denote the ordering type of $\AAA\cap]x, 1]$
for $x\leq\frac{1}{2}$ and the ordering type of $\AAA\cap[x, 1]$
for $x>\frac{1}{2}$.
If $x\geq y$, then $\alpha(x)\leq\alpha(y)$ because
$\AAA\cap]x, 1]\subseteq \AAA\cap]y, 1]$.
We will often use this inequality.

\begin{Lemma}
\label{lem:relat0}
$\alpha(\frac{1}{2})=\omega$.
\end{Lemma}

\noindent
{\bf Proof.}
$\AAA\cap]\frac{1}{2}, 1]$ consists of a single sequence
$1$, $2/3$, $3/5$, $\ldots$\cite{Freivalds}.
$\Box$

First, we prove lower bounds on the ordering type
of $\AAA$.
$l(p)$ is the largest ordinal $\alpha$ such that
there is an $\omega^{\alpha}$-sequence in $\AAA\cap]p, 1]$
which converges to $p$.
We define $l(p)=0$ if there is no such sequence
for any $\alpha$.

It is easy to see that $\alpha(p)\geq \omega^{l(p)}$.
However, there may be a large gap between these two ordinals.
For example, if $\AAA\cap]p, 1]$ has the ordering type $\omega^{\omega}+1$,
there is no infinite monotonic sequence converging to
$p$ and $l(p)=0$.
We use the function $l$ to prove lower bounds.

\begin{Lemma}
\label{lem:relat1}
\[l\left(\frac{p}{1+p}\right)\geq \alpha(p).\]
\end{Lemma}

\noindent
{\bf Proof.}
Transfinite induction over $p\in \AAA$.

\noindent
{\em Base Case.}
Let $p=1$. The ordering type of $\AAA\cap[1, 1]=\{1 \}$ is $1$.
The ordering type of $\AAA\cap]1/2, 1]$ is $\omega$ and $l(1/2)=1$.

\noindent
{\em Inductive Case.}
Consider two cases:
\begin{enumerate}
\item
$p$ is a successor element.

Let $p\in [\frac{1}{n}, \frac{1}{n-1}]$.
Let $r$ denote the element immediately preceding $p$.
We have $\alpha(p)=\alpha(r)+1$ because $p$ is the only element
of $\AAA\cap[p, 1]$ which does not belong to $\AAA\cap[r, 1]$.
By inductive assumption, $l(\frac{r}{1+r})\geq\alpha(r)$.

Consider the splitting of $[\frac{1}{n+1}, \frac{1}{n}]$
in the proof of Theorem \ref{Basic} (subsection \ref{subsub:split}).
In the first step, one of segments is $[\frac{p}{1+p}, \frac{r}{1+r}]$
because $[p, r]$ does not contain other elements of $\AAA$.
We consider the sequence $r_0$, $r_1$, $\ldots$
corresponding to $[\frac{p}{1+p}, \frac{r}{1+r}]$.

\begin{Claim}
\label{lem:relat2}
$l(r_i)\geq \alpha(r)$.
\end{Claim}

\noindent
{\bf Proof.}
By induction.

\noindent
{\em Base Case.}
Let $i=0$.
Then, $r_0=\frac{r}{1+r}$
and $l(r_0)=l(\frac{r}{1+r})\geq\alpha(r)$.

\noindent
{\em Inductive Case.}
We prove $l(r_{i+1})\geq l(r_i)$.
Then $l(r_{i+1})\geq \alpha(r)$ follows from $l(r_i)\geq\alpha(r)$.
We use

\begin{Claim}
\label{lem:omegainitial}
If a set is obtained from $\omega^{\alpha}$ by removing a proper
initial segment, it still has ordering type $\omega^{\alpha}$.
\end{Claim}

\noindent
{\bf Proof.}
If we remove a segment with ordering type $\beta$,
we obtain the set with ordering type $\omega^{\alpha}-\beta$
(Definition \ref{def:substr}).
We have $\omega^{\alpha}-\beta=\omega^{\alpha}$ for all
$\beta<\omega^{\alpha}$.
$\Box$

Let
\[f(x)=\frac{2}{1+\frac{1}{x}+\frac{1}{p}} .\]
$f(x)$ maps $x\in \AAA$ to the number
generated from $x$ and $p$ by rule \ref{Rule}(Lemma \ref{Formula}).
Let $x_0$ be such that $f(x_0)=r_i$.
The function $f$ maps $(x_0, r_i)$ to $(r_i, r_{i+1})$.
($r_{i+1}=f(r_i)$ by the definition of $r_{i+1}$.)

We take an $\omega^{\alpha(r)}$ sequence converging to $r_i$
and remove all $x<x_0$ from it.
The ordering type of the remaining sequence is still $\omega^{\alpha(r)}$
(Claim \ref{lem:omegainitial}).
$f$ maps it to a sequence
converging to $r_{i+1}$ and preserves the ordering.
Hence, $l(r_{i+1})\geq \alpha(r)$.
$\Box$

We take the union of $\omega^{\alpha(r)}$ sequences converging
to $r_0$, $r_1$, $\ldots$ and obtain a $\omega^{\alpha(r)+1}$
sequence converging to $\lim_{i\rightarrow\infty}r_i=\frac{p}{1+p}$.
Hence, $\l(\frac{p}{1+p})\geq \alpha(r)+1=\alpha(p)$.
\item
$p$ is a limit element.

Let $p_0, p_1, \ldots$ be a decreasing
sequence converging to $p$.
Then, $\alpha(p)=\lim_{i\rightarrow\infty}\alpha(p_i)$.

We take the union of $\omega^{\alpha(p_i)}$ sequences
converging to $\frac{p_i}{1+p_i}$.
It has the ordering type
\[\lim_{i\rightarrow\infty}{\omega^{\alpha(p_i)}}
=\omega^{\lim_{i\rightarrow\infty} \alpha(p_i)} = \omega^{\alpha(p)} \]
and converges to $\frac{p}{1+p}$.
Hence, $l(\frac{p}{1+p})\geq \alpha(p)$.
\end{enumerate}
$\Box$

\begin{Lemma}
\label{lem:relat1a}
$\alpha\left(\frac{p}{1+p}\right)\geq \omega^{\alpha(p)}$.
\end{Lemma}

\noindent
{\bf Proof.}
Follows from Lemma \ref{lem:relat1} and 
$\alpha(\frac{p}{1+p})\geq\omega^{l(\frac{p}{1+p})}$.
$\Box$

The upper bound proof is more complicated.
We prove a counterpart of Lemma \ref{lem:relat1a}.

\begin{Lemma}
\label{lem:upper1}
$\alpha\left(\frac{p}{1+p}\right)\leq \omega^{\alpha(p)}$.
\end{Lemma}

\noindent
{\bf Proof.}
Transfinite induction over $p\in \AAA$.

\noindent
{\em Base Case.}
Let $p=1$. The ordering type of $\AAA\cap[1, 1]$ is $1$ and
the ordering type of $\AAA\cap[1/2, 1]$ is $\omega$.

\noindent
{\em Inductive Case.}
Consider two cases:
\begin{enumerate}
\item
$p$ is a successor element.

Let $r$ be the element immediately preceding $p$.
We split the interval $[\frac{p}{1+p}, \frac{r}{1+r}]$
into subintervals $[r_1, r_0]$, $[r_2, r_1]$, $\ldots$,
as in section \ref{subsub:split}.

\begin{Claim}
\label{lem:upper2}
$\alpha(r_i)\leq c_i w^{\alpha(r)}$ for some $c_i\in\bbbn$.
\end{Claim}

\noindent
{\bf Proof.}
By induction.

\noindent
{\em Base Case.}
If $i=0$, $r_0=\frac{r}{1+r}$
and $\alpha(\frac{r}{1+r})\leq \omega^{\alpha(r)}$
by inductive assumption.

\noindent
{\em Inductive Case.}
Let $P$ be a $(r_{i+1}, r_{i+1})$-minimal set
(section \ref{subsub:xdminimal}).
Let $A(p_1, \ldots, p_s)$ denote
the set of all $x\in \AAA\cap]r_{i+1}, r_i]$
generated by applications of the rule \ref{Rule} to
$p'_1\in \AAA$, $\ldots$, $p'_s\in \AAA$ such that
$p_1\leq p'_1$, $\ldots$, $p_s\leq p'_s$.
$\alpha'(p_1, \ldots, p_s)$ denotes the ordering type
of $A(p_1, \ldots, p_s)$.

\begin{Claim}
\label{lem:upperpart1}
\[ \alpha(r_{i+1})\leq \alpha(r_i)(+)
 \sum_{\langle p_1, \ldots, p_s\rangle\in P}\alpha'(p_1, \ldots, p_s) .\]
\end{Claim}

\noindent
{\bf Proof.}
We have
\[ \AAA\cap]r_{i+1}, 1]=(\AAA\cap]r_i, 1])
\cup \bigcup_{\langle p_1, \ldots, p_s\rangle\in P}A(p_1, \ldots, p_s) .\]
By Lemma \ref{lem:ord1}, the ordering type of $\AAA\cap]r_{i+1}, 1]$
is less than or equal to the natural sum of the ordering types
of $\AAA\cap]r_i, 1]$ and all $A(p_1, \ldots, p_s)$.
$\Box$

Next, we bound each $\alpha'(p_1, \ldots, p_s)$.
We start with an auxiliary lemma.

\begin{Claim}
\label{lem:upperaux}
If $p\in \AAA$ follows from an application of the rule \ref{Rule}
to $p_1\in \AAA$, $\ldots$, $p_s\in \AAA$, then
\[ \alpha(p)\geq \alpha(p_1)(+)\ldots(+)\alpha(p_s) .\]
\end{Claim}

\noindent
{\bf Proof.}
Without the loss of generality, we assume that
$p_1\leq p_2\leq\ldots\leq p_s$.
Then, $\alpha(p_1)\geq \alpha(p_2)\geq\ldots\geq \alpha(p_s)$.
We prove the lemma by transfinite induction over $p_1$.

\noindent
{\em Base Case.}
$p_1$ is the maximum element, i.e. $p_1=1$.

Then, $p_1=\ldots=p_s=1$.
An application of the rule \ref{Rule} to $p_1$, $\ldots$, $p_s$
generates $p=s/(2s-1)$.
\[ \AAA\cap\left[\frac{s}{2s-1}, 1\right]=
\left\{\frac{s}{2s-1}, \frac{s-1}{2s-3}, \ldots,
\frac{2}{3}, 1\right\}. \]
The ordering type of this set is $s$, i.e. $\alpha(p)=s$.
On the other hand, $\alpha(p_1)=\ldots=\alpha(p_s)=1$ and
\[ \alpha(p_1)(+)\ldots(+)\alpha(p_s)=s .\]

\noindent
{\em Inductive Case.}
We have two possibilities:
\begin{enumerate}
\item
$p_1$ is a successor element.

$j$ denotes the maximum number such that $p_1=\ldots =p_j$.
Let
\[ p'_i=\cases{\mbox{predecessor of $p_1$}, &if $i\leq j$\cr
                p_i, &if $i>j$\cr}. \]
We have $\alpha(p_i)=\alpha(p'_i)+1$ for $i\leq j$ and
$\alpha(p_i)=\alpha(p'_i)$ for $i>j$.
Hence,
\[ \alpha(p_1)(+)\ldots(+)\alpha(p_s)=\]
\[ (\alpha(p'_1)+1)(+)\ldots(+)
(\alpha(p'_j)+1)(+)\alpha(p'_{j+1})(+)\ldots(+)\alpha(p'_s) =\]
\[ \alpha(p'_1)(+)\ldots(+)\alpha(p'_s)+j .\]
Let $x_0$ be the number generated by an application of the rule
\ref{Rule} to $p'_1$, $\ldots$, $p'_s$ and
$x_i$, for $i\in\{1, \ldots, j\}$, be
the number generated by an application of the rule
\ref{Rule} to $p_1$, $\ldots$, $p_i$, $p'_{i+1}$, $\ldots$, $p'_s$.
By inductive assumption,
\[ \alpha(x_0)\geq \alpha(p'_1)(+)\ldots(+)\alpha(p'_s).\]
We have $p_i< p'_i$ for $i\leq j$.
By Lemma \ref{Monotonicity}, $x_i< x_{i-1}$.
Hence,
\[ \alpha(x_i)\geq \alpha(x_{i-1})+1. \]
We have $p_i=p'_i$ for $i>j$.
Hence, $x_j=p$ and
\[ \alpha(p)=\alpha(x_j)\geq \alpha(x_0)+j \geq
\alpha(p'_1)(+)\ldots(+)\alpha(p'_s)+j =\]
\[\alpha(p_1)(+)\ldots(+)\alpha(p_s) .\]
\item
$p_1$ is a limit element.

Again, $j$ is the maximum number such that $p_1=\ldots =p_j$.
Let $p'_1, p'_2, \ldots$ be a monotonically decreasing sequence converging
to $p_1$. Without the loss of generality, we can assume that all
elements of $p'_1, p'_2, \ldots$ are less than or equal to $p_{j+1}$.
(Otherwise, just remove the elements that are larger than $p_{j+1}$ 
and use the sequence consisting of remaining elements.)

Let $x_i$ be the number generated by an application of the
rule \ref{Rule} to $p'_i$, $\ldots$, $p'_i$, $p_{j+1}$, $\ldots$, $p_s$.
By inductive assumption,
\begin{equation}
\label{eq:4.1}
\alpha(x_i)\geq\underbrace{\alpha(p'_i)(+)\ldots(+)\alpha(p'_i)}_{\mbox{
\scriptsize $j$ times}}(+)\alpha(p_{j+1})(+)\ldots(+)\alpha(p_s) .
\end{equation}
We have $p_1=\ldots=p_j=\lim_{i\rightarrow\infty}p'_i$.
By Lemma \ref{Limits}, $p=\lim_{i\rightarrow\infty}x_i$.
Hence, if we take $i\rightarrow\infty$ in (\ref{eq:4.1})
and apply the fact that $(+)$ is continuous, we get
\[ \alpha(p)\geq \alpha(p_1)(+)\ldots(+)\alpha(p_s) .\]
\end{enumerate}
$\Box$

\begin{Claim}
\label{lem:upperpart2}
Let $\langle p_1, \ldots, p_s\rangle\in P$.
Then
\[ \alpha'(p_1, \ldots, p_s)\leq \const \omega^{\alpha(r)}.\]
\end{Claim}

\noindent
{\bf Proof.}
Lemma \ref{lem:ord2} implies that
$\alpha'(p_1, \ldots, p_s)$ is at most the natural product of
$\alpha(p_1)$, $\ldots$, $\alpha(p_s)$.
Let $\alpha_j$ be the largest ordinal such that
$\alpha(p_j)\geq\omega^{\alpha_j}$.
Then, $\alpha(p_j)\leq c_j\omega^{\alpha_j}$.
(If there is no such $c_j$, then
$\alpha(p_j)\leq \lim_{c\rightarrow\infty} 
c\omega^{\alpha_j}=\omega^{\alpha_j+1}$
and $\alpha_j$ is not the largest ordinal with this property.)
Hence,
\[ \alpha'(p_1, \ldots, p_s)\leq
c_1 \omega^{\alpha_1} (\cdot) c_2 \omega^{\alpha_2}
 \ldots c_s \omega^{\alpha_s}=
(c_1 c_2\ldots c_s) \omega^{\alpha_1(+)\alpha_2(+)\ldots(+)\alpha_s}.\]
Let $p'_j$ be such that $p'_j\in \AAA$ and $\alpha(p'_j)=\alpha_j$.
We have $\alpha(p'_j)=\alpha_j\leq \alpha(r)$ because 
\[ \omega^{\alpha_j} \leq \alpha(p_j) \leq
\alpha(r_i)\leq \const \omega^{\alpha(r)}< \omega^{\alpha(r)+1} ,\]
where $\alpha(p_j)\leq \alpha(r_i)$ follows from 
$p_j\geq r_i$.
$\alpha(p'_j)=\alpha_j\leq \alpha(r)$ implies $p'_j\geq r$.
Therefore, both Lemma \ref{lem:relat1} and
Lemma \ref{lem:upper1} are true for $p=p'_j$.
This means that $\alpha(\frac{p'_j}{1+p'_j})=\omega^{\alpha_j}$.
Hence, $\frac{p'_j}{1+p'_j}\geq p_j$ because
$\alpha(p_j)\geq \omega^{\alpha_j}$.

Let $p'$ be the number generated by an application of the rule \ref{Rule}
to $p'_1$, $\ldots$, $p'_s$.
By Lemma \ref{lem:rule}, $\frac{p'}{1+p'}$ is generated by
an application of the rule \ref{Rule}
to $\frac{p'_1}{1+p'_1}$, $\ldots$, $\frac{p'_s}{1+p'_s}$.
$\frac{p'}{1+p'}$ is greater than
or equal to the number generated by an application
of rule \ref{Rule} to $p_1$, $\ldots$, $p_s$
because $\frac{p'_j}{1+p'_j}\geq p_j$.
This number is at least $r_{i+1}$ because the tuple
$\langle p_1, \ldots, p_s\rangle$ belongs to
the $(r_{i+1}, r_{i+1})$-allowed set $P$.
Hence, $\frac{p'}{1+p'}\geq r_{i+1}$.
We have $\frac{p'}{1+p'}\geq \frac{r}{1+r}$ because
$[\frac{p}{1+p}, \frac{r}{1+r}]$ does not contain any points
of type $\frac{p'}{1+p'}$ with $p'\in \AAA$.
This implies $p'\geq r$.

By Claim \ref{lem:upperaux}, 
\[ \alpha(p')\geq\alpha(p_1)(+)\ldots(+)\alpha(p_s) .\]
This implies 
\[ \alpha'(p_1, \ldots, p_s)\leq (c_1\ldots c_s) 
\omega^{\alpha(p_1)(+)\ldots(+)\alpha(p_s)}\leq\]
\[ (c_1\ldots c_s) \omega^{\alpha(p')}\leq 
(c_1\ldots c_s) \omega^{\alpha(r)} .\]
$\Box$

Now, we are ready to finish the proof of Claim \ref{lem:upper2}.
By Claim \ref{lem:upperpart1}, $\alpha(r_{i+1})$ is less than or
equal to the natural sum of $\alpha(r_i)$ and
$\alpha'(p_1, \ldots, p_s)$.
We have
$\alpha(r_i)\leq \const \omega^{\alpha(r)}$ by inductive
assumption and
\[ \alpha'(p_1, \ldots, p_s)\leq const \omega^{\alpha(r)}\]
by Claim \ref{lem:upperpart2}.
Hence, the natural sum of these ordinals is at most 
$\const \omega^{\alpha(r)}$, too.
$\Box$
\[ \alpha\left(\frac{p}{1+p}\right)
=\lim_{i\rightarrow\infty}\alpha(r_i) \leq
 \lim_{i\rightarrow\infty}c_i \omega^{\alpha(r)} \leq
 \lim_{i\rightarrow\infty}\omega \cdot\omega^{\alpha(r)}=
 \omega^{\alpha(r)+1}=\omega^{\alpha(p)} .\]
\item
$p$ is a limit element.

Let $p_0, p_1, \ldots$ be a decreasing
sequence converging to $p$.
By inductive assumption,
$\alpha(\frac{p_i}{1+p_i})\leq \omega^{\alpha(p_i)}$.
We have
\[\alpha\left(\frac{p}{1+p}\right)=
\lim_{i\rightarrow\infty}\alpha\left(\frac{p_i}{1+p_i}\right)
\leq \lim_{i\rightarrow\infty}\omega^{\alpha(p_i)}
=\omega^{\lim_{i\rightarrow\infty} \alpha(p_i)} = \omega^{\alpha(p)}. \]
$\Box$
\end{enumerate}

\begin{Lemma}
\label{lem:relat1b}
$\alpha(\frac{p}{1+p})=\omega^{\alpha(p)}$.
\end{Lemma}

\noindent
{\bf Proof.}
Follows from Lemmas \ref{lem:relat1a} and \ref{lem:upper1}.
$\Box$

\begin{Theorem}
The ordering type of $\AAA$ is at least $\epsilon_0$.
\end{Theorem}

\noindent
{\bf Proof.}
The ordering type of $\AAA\cap(\frac{1}{2}, 1]$ is
$\omega$ (Lemma \ref{lem:relat0}).
The ordering type of $\AAA\cap(\frac{1}{3}, 1]$ is
$\omega^{\omega}$ (Lemma \ref{lem:relat1b}
with $p=1/2$), the ordering type of $\AAA\cap(\frac{1}{4}, 1]$
$\omega^{\omega^{\omega}}$ and so on.

The ordering type of $\AAA$ is the limit of this sequence, i.e.
\[ \epsilon_0=\lim (\omega, \omega^{\omega}, \omega^{\omega^{\omega}},
  \omega^{\omega^{\omega^{\omega}}}, \ldots ) .\]
$\Box$

It is known that the ordinal $\epsilon_0$ expresses the
set of all expressions possible in first-order arithmetic.
%
We see that $\PFIN$, a very simple learning criterion,
generates a very complex probability hierarchy.

The table below shows how the complexity of the hierarchy 
increases.
All results in this table can be obtained using
Lemma \ref{lem:relat1b}.

\begin{tabular}{|c||c|}
\hline
Interval & Ordering type of the probability hierarchy \\
\hline
\hline
$[\frac{1}{2}, 1]$ & $\omega$ \\
\hline
$[\frac{4}{9}, 1]$ & $2\omega$ \\
\hline
$[\frac{3}{7}, 1]$ & $3\omega$ \\
\hline
$[\frac{2}{5}, 1]$ & $\omega^2$ \\
\hline
$[\frac{3}{8}, 1]$ & $\omega^3$ \\
\hline
$[\frac{1}{3}, 1]$ & $\omega^{\omega}$ \\
\hline
$[\frac{1}{4}, 1]$ & $\omega^{\omega^{\omega}}$ \\
\hline
$[0, 1]$ & $\epsilon_0$ \\
\hline
\end{tabular}

It shows that the known part of hierarchy ($[\frac{3}{7}, 1]$)
is very simple compared to the entire hierarchy.

{\bf Notes.}
The points of the probability hierarchy in the intervals 
$[\frac{1}{2}, 1], [\frac{4}{9}, 1]$ and $[\frac{3}{7}, 1]$
were explicitly described in \cite{Freivalds}, \cite{DKV92a}
and \cite{DK93}, respectively.

In \cite{DK93}, an $\omega^2$ sequence of points converging
to $\frac{2}{5}$ was presented and it was conjectured that 
this sequence forms the backbone of the learning capabilities 
in the interval $[\frac{2}{5}, 1]$.

\section{Probabilistic versus team learning}

For $\EX$-identification, there is a precise correspondence
between probabilistic and team learners
(Pitt's connection\cite{Pitt}).
Any probabilistic learner can be simulated by any team
with the ratio of successful machines
equal to the probability of success for the probabilistic learner.

However, the situation is more complicated
for finite learning ($\FIN$ and $\PFIN$).
Here, the learning power of a team depends not
only on the ratio of successful machines.
Team size is also important.

\begin{Theorem}
\cite{Vel,JSV}
$[1, 2]\PFIN \subset [2, 4]\PFIN$.
\end{Theorem}

So, a team of 4 learning machines where 2 machines are required 
to be successful has more learning power than team
of 2 learning machines where 1 must succeed.
However, in both teams the ratio of successful machines
to all machines is the same($\frac{1}{2}$).

This phenomena is called {\em redundancy}.
Various redundancy types have been discovered for various
ratios of successful machines \cite{DKV91,DK93,JSV}.
The theorem below is the example of infinite redundancy\cite{DK93,DKV91}.

\begin{Theorem}
\label{InfiniteRedundancy}
\cite{DK93}
It $k \bmod 3\neq 0$, then 
\[ [2k, 5k]\PFIN \subset [8k, 20k]\PFIN .\]
In particular, 
\[ [2, 5]\PFIN \subset [8, 20]\PFIN \subset [32, 80]\PFIN\subset\ldots.\] 
\end{Theorem}

So, for the ratio of successful machines 2/5 there are infinitely many
different team sizes with different learning power.

However, even for $\PFIN$, any probabilistic machine
can be simulated by a team with the same ratio of success,
if we choose the team size carefully.
A simple corollary of Theorem \ref{MainSimulation} is

\begin{Corollary}
\label{ProbabilisticTeam}
If $p, q\in\bbbn^{+}$, then there exists $k$ such that
\[ \PFIN\langle \frac{p}{q}\rangle = [pk, qk]\PFIN .\]
\end{Corollary}

This shows that probabilistic $\PFIN$-learning and team $\PFIN$-learning
are of the same power.

\begin{Corollary}
If $p, q\in\bbbn^{+}$, then there exists $k$ such that
\[ [pl, ql]\PFIN \subseteq [pk, qk]\PFIN \]
for any $l\in\bbbn^{+}$.
\end{Corollary}

{\bf Proof.}
The team of $ql$ machines can be simulated by single
probabilistic machine which equiprobably chooses
one of machines in team and simulates it.
Hence, Corollary \ref{ProbabilisticTeam}
implies that
\[ [pl, ql]\PFIN \subseteq PFIN\langle \frac{p}{q}\rangle = [pk, qk]\PFIN .\]
$\Box$

So, we see that redundancy structures can be very complicated
but always there is the "best" team size such that team of this size
can simulate any other team with the same ratio of
successful machines.
It exists even if there are infinitely many team sizes 
with different learning power (like for ratio 2/5, Theorem 
\ref{InfiniteRedundancy}).

\section{Conclusion}
\label{sec:conclusion}

We have investigated the structure of probability hierarchy
for $\PFIN$-type learning.
Instead of trying to determine the exact points 
at which the learning capabilities change,
we focused on the structural properties of the hierarchy.

We have developed a universal diagonalization algorithm
(Theorem \ref{Part1}) and a universal simulation
algorithm (Theorem \ref{MainSimulation}).
These algorithms are very general forms of diagonalization
and simulation arguments used for probabilistic $\PFIN$
\cite{DK93,DKV92a}.

Universal diagonalization theorem gives the method that
can be used to obtain any possible diagonalization
for probabilistic $\PFIN$.
Universal simulation algorithm can be used for
any possible simulation.

These two results together give us a recursive description
of the set of points $\AAA$ at which the learning capabilities
are different.

This set is well-ordered in decreasing ordering.
(This property is essential to the proof of Theorem
\ref{MainSimulation}.)
Its structure is quite complicated.
Namely, its ordering type is $\epsilon_0$,
the ordering-type of the set of all expressions 
possible in first-order arithmetic.

It shows the huge complexity of the probabilistic $\PFIN$-hierarchy
and explains why it is so difficult to find the points at which
the learning capabilities are different.

A simple corollary of our results is that
the probabilistic and team $\PFIN$-type learning is of the same
power, i.e. any probabilistic learning machine
can be simulated by a team with the same success ratio.

Several open problems remain:
\begin{enumerate}
\item
Unrestricted finite learning($\FIN$).

The major open problem is the generalization
of our results for other learning paradigms such as
(non-Popperian) $\FIN$-type learning and
language learning in the limit.

Theorem \ref{Part1} can be proved for 
(nonPopperian) $\FIN$-type learning, too.
Hence, if 
\[\PFIN\langle p_1\rangle\neq \PFIN\langle p_2\rangle,\] 
then
\[ \FIN\langle p_1\rangle\neq \FIN\langle p_2\rangle .\]
So, the probability hierarchy of $\FIN$ is at least as complicated
as the probability hierarchy of $\PFIN$.
It is even more complicated because it is known\cite{DKV92,DKV92a} that
\[\FIN\langle 24/49 \rangle\subset \FIN\langle 1/2\rangle\]
but 
\[\PFIN\langle 24/49 \rangle= \PFIN\langle 1/2\rangle.\]

The simulation techniques for $\FIN$ are much more complicated than
simulation techniques for $\PFIN$.
However, we hope that some combination of our methods
and other ideas (e.g. \cite{DKV92,DK97}) can help
to identify the set of all possible diagonalization
methods for $\FIN$ and to prove that no other
diagonalization methods exists (i.e. to construct
universal simulation for $\FIN$).

A step in that direction was made in \cite{Apsitis97}
by proving that $\FIN$-hierarchy is well-ordered and
recursively enumerable.
It still remains open whether it is decidable.
The proof technique in \cite{Apsitis97} is different from
ours and uses capability trees\cite{DK97}.
\item
Probabilistic language learning.

The probability hierarchy of language learning
in the limit\cite{JS93a}
has some similarities to $\FIN$ and $\PFIN$-hierarchies.

It is an interesting open problem whether 
some analogues of our results can
be obtained for language learning in the limit.
\item
What is the computational complexity of decision algorithms
for $\PFIN$-hierarchy?
\item
How dense is the probability hierarchy?

Can we prove the result of the following type:

If $p_1, p_2\in [\frac{1}{n+1},\frac{1}{n}]$ and $|p_1-p_2|<(1/2)^n$,
then 
\[ \PFIN(p_1)\neq \PFIN(p_2) ?\]
\end{enumerate}
Other properties of the whole hierarchy can be studied, too.

{\bf Acknowledgments.}
I would like to thank the referee for the valuable comments that
helped to improve this paper.
This research was done while the author was at the University of Latvia and
supported by Latvian Science Council Grants
No.93.599 and No.96.0268 and fellowship
"SWH izgl\=\i t\=\i bai, zin\=atnei un kult\=urai".
Extended abstract of this paper appeared on the 9th Conference on
Computational Learning Theory, Desenzano del Garda, Italy, 1996. 

\end{document}